\documentclass[10pt,twocolumn,letterpaper]{article}

\usepackage{cvpr}              
\definecolor{cvprblue}{rgb}{0.21,0.49,0.74}
\usepackage[pagebackref,breaklinks,colorlinks,allcolors=cvprblue]{hyperref}

\def\paperID{} 
\def\confName{CVPR}
\def\confYear{2026}

\title{Seeing is Improving: Visual Feedback for Iterative Text Layout Refinement}

\author{
Junrong Guo\textsuperscript{1}, Shancheng Fang\textsuperscript{2}\thanks{Corresponding author.}, Yadong Qu\textsuperscript{1}, Hongtao Xie\textsuperscript{1}\\
\textsuperscript{1}University of Science and Technology of China\quad
\textsuperscript{2}Shenzhen University\\
{\tt\small godrong@mail.ustc.edu.cn, fangsc@szu.edu.cn, qqqyd@mail.ustc.edu.cn, htxie@ustc.edu.cn}
}

\begin{document}
\maketitle
\begin{abstract}
Recent advances in Multimodal Large Language Models (MLLMs) have enabled automated generation of structured layouts from natural language descriptions. Existing methods typically follow a code-only paradigm that generates code to represent layouts, which are then rendered by graphic engines to produce final images. However, they are blind to the rendered visual outcome, making it difficult to guarantee readability and aesthetics. In this paper, we identify visual feedback as a critical factor in layout generation and propose Visual Feedback Layout Model (VFLM), a self-improving framework that leverages visual feedback iterative refinement. 
VFLM is capable of performing adaptive reflective generation, which leverages visual information to reflect on previous issues and iteratively generates outputs until satisfactory quality is achieved.
It is achieved through reinforcement learning with a visually grounded reward model that incorporates OCR accuracy. By rewarding only the final generated outcome, we can effectively stimulate the model's iterative and reflective generative capabilities.
Experiments across multiple benchmarks show that VFLM consistently outperforms advanced MLLMs, existing layout models, and code-only baselines, establishing visual feedback as critical for design-oriented MLLMs. Our code and data are available at \url{https://github.com/FolSpark/VFLM}.
\end{abstract}      
\section{Introduction}
\label{sec:introduction}

The emergence of Large Language Models (LLMs)~\cite{achiam2023gpt,deepseekai2025deepseekr1incentivizingreasoningcapability,yang2025qwen3} and Multimodal Large Language Models (MLLMs)~\cite{hurst2024gpt,o3,qwen2.5-vl,seed2025seed1_5vl} has unlocked new possibilities for automated content generation, particularly for structured visual layouts. By generating structured representations (e.g., SVG code, custom JSON)~\cite{feng2024layoutgpt,yang2024posterllava,graphist2023hlg,qu2025igd} that define the position, size, and style of each element~\cite{jia2023cole,inoue2024opencole}, LLMs can translate natural language descriptions directly into complex designs such as typographic posters, social media graphics, and documents. MLLMs further enhance this by leveraging cross-modal understanding, allowing generation to be conditioned on both text and visual inputs.
\begin{figure}[tbp]
    \centering
    \includegraphics[width=1\linewidth]{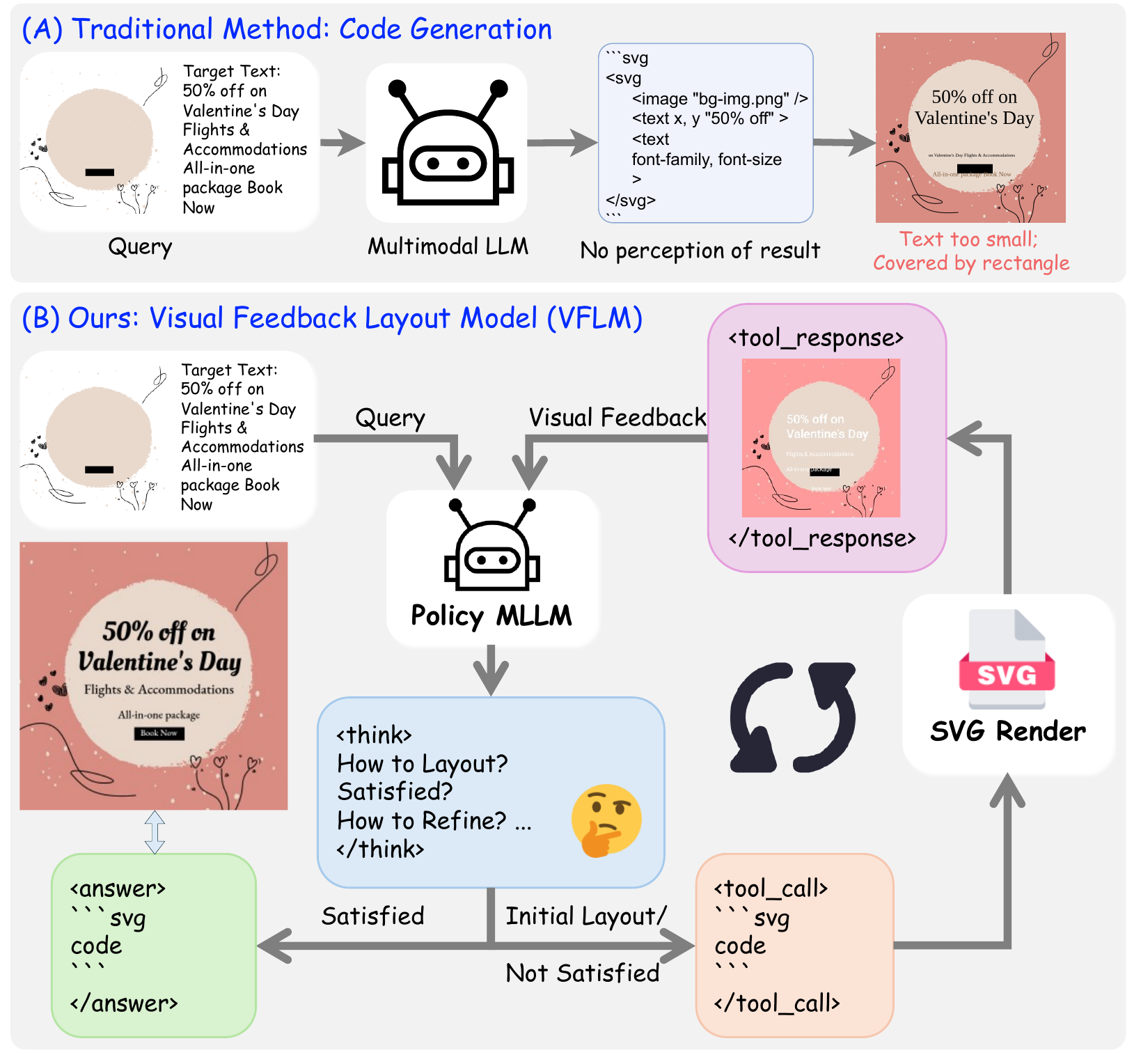}
    \caption{Method comparison. Traditional methods only generate through text code and have no perception of the generated results; our Visual Feedback method can iteratively reflect on the generated result and optimize the layout effect.}
    \label{fig:method_compration}
    \vspace*{-1.5\baselineskip}
\end{figure}

However, existing methods are limited to the code-only paradigm and can only generate codes for layout without considering visual feedback. For instance, COLE~\cite{jia2023cole} and OpenCOLE~\cite{inoue2024opencole} leverage LLMs to generate typography JSON files, while LGGPT~\cite{lggpt2025zhang} produces customized layout output formats, which are then composed into the final images by a graphic renderer. 
As shown in \cref{fig:method_compration}(A), while these models can produce layout structures that meet formal specifications, they remain unable to perceive the visual outcomes of their designs. Effective text layout design, however, depends on visual properties such as aesthetic balance, readability, and image–text coherence, which cannot be fully encoded by programmatic rules. Consequently, a model may generate syntactically valid SVG layouts that still suffer from overlapping elements, poor contrast, or misaligned components.

Recent advances in LLMs~\cite{deepseekai2025deepseekr1incentivizingreasoningcapability,jaech2024openai,o3,gandhi2025cognitive} research demonstrate that reflection, backtracking, and self-validation mechanisms can substantially improve performance on complex reasoning tasks. Moreover, Reinforcement Learning (RL)~\cite{NEURIPS2022_rlhf,schulman2017proximal,shao2024deepseekmath} techniques have proven effective in activating the reflective reasoning capabilities of LLMs. This motivates our core research question: Can such reflection capabilities be transferred to text layout generation to overcome the visual perception gap in existing approaches?
We argue that the solution lies in incorporating \textbf{visual feedback} into the text layout generation process, leveraging MLLMs' inherent cross-modal understanding capabilities. As illustrated in \cref{fig:method_compration}, models should not only generate layout code but also perceive the rendered results to evaluate quality, diagnose visual issues, and devise optimization strategies through iterative refinement.

In this paper, we propose \textbf{V}isual \textbf{F}eedback \textbf{L}ayout \textbf{M}odel (VFLM), a novel self-improving framework for text layout generation that establishes a closed-loop process guided by RL. As shown in \cref{fig:method_compration}, VFLM first generates initial SVG layout code, which is rendered into a visual image. This rendered image is fed back to the same model for visual inspection and reflection. If issues are identified or the layout is unsatisfactory, VFLM generates revised code and repeats the process, creating a continuous loop of ``\textit{generation, rendering, reflection and refinement}'' until a satisfactory layout is achieved. 

VFLM utilizes a two-stage training pipeline. Firstly, we construct a multi-stage generation-reflection-refinement trajectory dataset by distilling advanced MLLMs, followed by Supervised Fine-Tuning (SFT) to endow model the ability of iterative generation. Secondly, we employ RL to enhance the model's reflective capabilities. We utilize a reward model trained to evaluate layout quality holistically and incorporate text accuracy via Optical Character Recognition (OCR). While recent research on Agentic RL research~\cite{singh2025agentic,dong2025agentic,li2025deepagent} often relies on meticulously designed, complex, process-oriented reward mechanisms to guide each intermediate step, our analysis reveals this approach can lead to ``reward hacking''. 
We demonstrate, counter-intuitively, that a simple outcome-based reward, which only assesses the final generated layout, is significantly more effective. This simpler signal compels the VFLM to leverage its inherent visual understanding to holistically balance initial generation quality and iterative refinement, enabling robust and stable performance improvements through visual feedback.

We validate this framework through extensive experiments based on Qwen2.5-VL-7B model~\cite{qwen2.5-vl} for the task of laying out target text on background images. Both quantitative and qualitative evaluations show VFLM significantly outperforms state-of-the-art layout generation approaches, advanced MLLMs, and image editing models. Furthermore, our ablation studies, which provide fair comparisons against multiple code-only generation methods, confirm the significant and critical advantage of our VFLM. This superiority effectively validates visual feedback as a crucial component in generative text layout, establishing a practical framework for developing self-improving, MLLM-based design agents. 

This work's contributions can be summarized as follows: 
\begin{itemize}
    \item We propose VFLM, a self-improving framework that, to our knowledge, is the first to apply visual feedback to layout generation. It equips MLLMs with a ``generation, rendering, reflection, refinement'' cycle, using visual feedback to overcome the limitations of code-only methods.
    \item We design a two-stage SFT+RL training method, including a novel reward model for text layout, that successfully activates the model's iterative refinement capabilities. Extensive experiments validate that our approach consistently outperforms strong existing methods.
    \item We demonstrate a key finding that simple, outcome-based rewards are more effective and robust for activating the model's self-improvement capabilities than complex, process-oriented reward functions.
\end{itemize}

\section{Related Work}
\subsection{Multimodal Large Language Model}
Recent progress in Multimodal Large Language Models (MLLMs) is driven by integrating pretrained vision encoders ~\cite{radford2021learning,zhai2023sigmoid} with LLMs. The two modalities are typically aligned via lightweight projectors or Q-Former ~\cite{li2023blip2} structures, a paradigm that has spurred a suite of powerful models. This includes open-source series like LLaVA ~\cite{liu2023llava,liu2023improvedllava,liu2024llavanext}, Qwen-VL ~\cite{qwen-vl, qwen2-vl, qwen2.5-vl}, and Intern-VL ~\cite{chen2024internvl, gao2024mini-internvl, zhu2025internvl3, wang2025internvl3_5}, as well as large-scale proprietary systems such as GPT-4o ~\cite{hurst2024gpt}, Gemini ~\cite{team2023gemini}, and Claude ~\cite{anthropic2025claude}, which continue to advance the state of the art through massive scaling and enhanced reasoning techniques ~\cite{wei2022chain,zhang2025survey}.

\subsection{Graphic Layout Generation}
Graphic layout generation has evolved from early Transformer-based architectures~\cite{zhou2022composition,lin2023autoposter} to methods centered on LLMs and MLLMs. 
Many approaches prompts LLMs to output structured formats like SVG~\cite{wang2025svgen}, HTML~\cite{seol2024posterllama}, or JSON~\cite{lin2023layoutprompter,jia2023cole,inoue2024opencole,textlap,hsu2025postero}. Others employ MLLMs to decompose the design process into ordered layers or sub-tasks~\cite{graphist2023hlg,lin2025elements,jia2023cole,qu2025igd}, often coordinating with multi-modal inputs~\cite{yang2024posterllava,inoue2023document}. 
However, these methods operate in an open-loop, lacking visual perception of the rendered output. 
While some recent pipelines incorporate visual feedback, they typically rely on non-trainable, inference-only strategies with external advisors~\cite{yang2024idea2img,zhang2024vascar}, or decouple generation and reflection into separately trained models~\cite{jia2023cole}. 
In contrast, our work introduces visual feedback via RL, unifying these capabilities into a single, self-evolving policy with learnable autonomous judgment.

\subsection{Reinforcement Learning}
Reinforcement learning is a cornerstone for aligning LLMs with human preferences, standardized by the RLHF~\cite{NEURIPS2022_rlhf} pipeline which typically uses Proximal Policy Optimization (PPO)~\cite{schulman2017proximal}. To mitigate the instability and high cost associated with PPO, recent alternatives like Direct Preference Optimization (DPO)~\cite{rafailov2023direct} and Group Relative Policy Optimization (GRPO)~\cite{shao2024deepseekmath} offer more direct and efficient optimization strategies.
This alignment paradigm extends naturally to multimodal settings to improve visual grounding and reasoning. Works such as Vision-R1~\cite{huang2025vision}, R1VL~\cite{zhang2025r1}, and DeepEyes~\cite{zheng2025deepeyesincentivizingthinkingimages} adapt RL to MLLMs by incorporating multimodal rewards, chain-of-thought signals, and specialized replay mechanisms, demonstrating the power of RL in enhancing multimodal alignment and capability.

\section{Method}
\label{sec:method}
\begin{figure*}[t]
    \centering
    \includegraphics[width=1\linewidth]{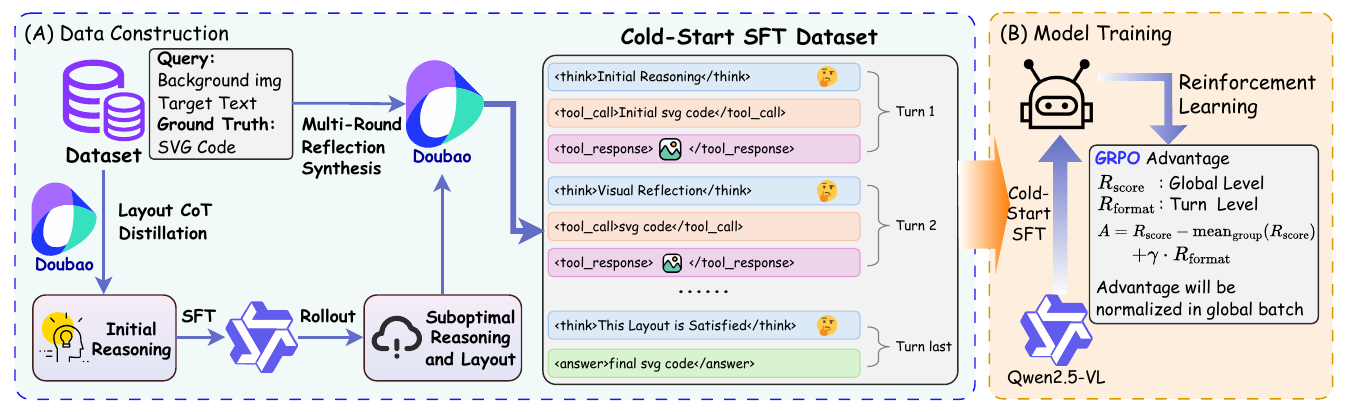}
    \caption{Pipeline of Visual Feedback Method, including Data Construction for Cold-Start SFT and RL with Modified Advantages.}
    \label{fig:data_and_train}
    \vspace*{-\baselineskip}
\end{figure*}
This work aims to develop a self-improving VFLM for text layout generation that optimizes outputs through visual feedback. We achieve this with a two-stage training framework: (1) Cold-Start SFT to instill basic iterative generation and reflection capabilities, and (2) Reinforcement Learning to enhance performance using vision-based reward signals.

\subsection{Visual Feedback Layout Model}
\label{subsec:task_formulation}
As illustrated in \cref{fig:method_compration}, VFLM establishes a multi-round interaction mechanism between VFLM and a rendering environment. As shown in \Cref{alg:reflection_process}, given a background image and target text, VFLM executes an iterative \textit{generation, rendering, reflection, refinement} cycle:

\begin{enumerate}
    \item \textbf{Initial Generation:} The VFLM first analyzes the input through reasoning, then generates initial layout code through a structured tool call.
    \item \textbf{Rendering:} The rendering tool converts the SVG code into a visual image and feeds it back to the same VFLM.
    \item \textbf{Visual Reflection:} The VFLM examines the rendered layout visual image through reasoning to evaluate whether the quality is satisfactory to it. 
    \item \textbf{Iterative Refinement:} If unsatisfied, the VFLM reasons about necessary modifications and a revised layout code is generated. The process repeats until the model determines satisfaction with the layout quality.
\end{enumerate}

\begin{algorithm}[htbp]
\caption{Visual Feedback for Text Layout Self-Improvement}
\label{alg:reflection_process}
\begin{algorithmic}[1]
\Require Background Image $I_b$, Target Text $T$, max iterations $N_{\text{max}}$
\Ensure Final layout code $S$

\State $S \gets $ VFLM generates initial layout via tool call based on $ I_b \text{ and } T$
\State $I_{\text{rendered}} \gets \text{Render}(S)$ \Comment{Tool Call}

\For{$i = 1 \to N_{\text{max}}$}
    \State $\text{Reflection} \gets \text{VFLM}(\text{reason} | I_{\text{rendered}})$
    
    \If{$\text{Reflection indicates satisfaction}$}
        \State \Return $S$ as the final result
    \EndIf
    
    \State $S \gets \text{VFLM generates revised layout}$
    \State $I_{\text{rendered}} \gets \text{Render}(S)$ \Comment{Tool Call}
\EndFor

\State \Return $S$
\end{algorithmic}
\end{algorithm}

\subsection{Cold-Start Supervised Fine-Tuning}
\label{subsec:sft}

The cold-start SFT stage enables the model to acquire iterative reflection capabilities, and tool usage specifications through distillation from an advanced teacher MLLM.

\noindent \textbf{Data Construction}
Due to the absence of natural multi-round reflection data, we employ Doubao-Seed-1.6~\cite{doubao1.6} as a teacher model for data synthesis. As illustrated in \cref{fig:data_and_train}(A), the systhesis pipeline is divided into the following stages:
\begin{enumerate}
    \item \textbf{Initial reasoning synthesis:} First, The teacher model is prompted with background images and ground-truth layouts to generate the reasoning processes for SVG code generation.
    \item \textbf{Suboptimal layout generation:} This distilled reasoning data is used to fine-tune Qwen2.5-VL-7B. The inference outputs from this model are then collected to serve as the suboptimal initial attempts for the next stage. 
    \item \textbf{Multi-round reflection synthesis:} The suboptimal attempts (from Stage 2) are paired with their ground-truth layouts and fed back to the teacher model. The teacher is then instructed to perform iterative reflection and modification to reach the ground-truth solution. This process simulates realistic human design refinement and generates complete multi-round reflection trajectories. 
    \item \textbf{Data combination:} Finally, the synthesized data from the above stages are combined and organized using structured tags: intermediate rounds use \texttt{<think>} and \texttt{<tool\_call>}, while the final round uses \texttt{<think>} and \texttt{<answer>} (See supplementary materials for specific data samples.). 
\end{enumerate}

\noindent \textbf{Training Objective}
We fine-tune Qwen2.5-VL-7B using causal language modeling on the synthesized multi-round reflection trajectories data. To prevent the model from learning suboptimal outputs, the loss for initial responses in improvement sequences is masked, ensuring the model learns from the correction process rather than initial errors.

\subsection{Iterative Reflection Reinforcement Learning}
\label{subsec:rl}

\subsubsection{RL Algorithm}
We adopt the Group Relative Policy Optimization (GRPO)~\cite{shao2024deepseekmath} algorithm for RL and make certain improvements to the advantage function. Compared with traditional policy optimization methods, GRPO performs policy gradient optimization within sample groups, enabling the model to learn in the direction of maximizing rewards. The optimization objectives of GRPO are as follows:
\vspace{-0.2\baselineskip}
\begin{equation}
  \resizebox{\linewidth}{!}{$%
    \begin{aligned}
      \mathcal{J}_{\text{GRPO}}(\theta)
      & =\mathbb{E}\!\left[q \sim P(Q),\{o_i\}_{i=1}^{G} \sim \pi_{\theta_{\text{old}}}(O \mid q)\right] \\
      &\quad \frac{1}{G}\sum_{i=1}^{G}\frac{1}{|o_i|}\sum_{t=1}^{|o_i|}\!\Bigl\{
          \min\!\Bigl[
            \frac{\pi_{\theta}(o_{i,t}\!\mid\! q, o_{i,<t})}{\pi_{\theta_{\text{old}}}(o_{i,t}\!\mid\! q, o_{i,<t})}
            A_{i,t}, \\
      & 
            \operatorname{clip}\!\Bigl(
              \frac{\pi_{\theta}(o_{i,t}\!\mid\! q, o_{i,<t})}{\pi_{\theta_{\text{old}}}(o_{i,t}\!\mid\! q, o_{i,<t})},
              1-\varepsilon, 1+\varepsilon
            \Bigr) A_{i,t}
          \Bigr] -\beta\,\mathbb{D}_{\mathrm{KL}}\!\bigl[\pi_{\theta}\,\|\,\pi_{\mathrm{ref}}\bigr]\Bigr\},
    \end{aligned}$}%
\end{equation}
where $\epsilon$ and $\beta$ are the clipping hyperparameters and the KL divergence penalty coefficient, respectively.

For the reward function, we design a three-component reward function to score the layout effect: (1) $R_\text{layout}$(\cref{subsec:rm_train}), a specialized reward model trained to evaluate overall layout quality; (2) $R_\text{ocr}$, which evaluates text accuracy based on OCR results from the rendered layout; and (3) $R_\text{svg}$, which measures text accuracy at the code level by comparing the extracted SVG strings with the target text.
The total reward is the weighted sum of the three components:
\vspace{-0.5\baselineskip}
\begin{equation}
    R_\text{score} = R_{\text{layout}} + \alpha \cdot \left( R_{\text{ocr}} + R_{\text{svg}}\right),
\end{equation}
where $\alpha$ balance aesthetic quality against functional accuracy.
In addition, we incorporate a format reward to constrain the output format of the model:
\vspace{-0.5\baselineskip}
\begin{equation}
    R_{\text{format}} =
    \begin{cases}
    1.0, & \text{if format is correct}, \\
    -1.0, & \text{if format is incorrect}.
    \end{cases}
    \label{eq:format_reward}
\end{equation}
For the advantage function, since the multi-round nature of our approach, the format rewards are applied separately in each round, resulting in inconsistent rewards between rounds for a sequence. Referring to REINFORCE++~\cite{hu2025reinforce++}, we use the mean value of $R_\text{score}$ within a group as baseline to reshape the reward. The format rewards are then incorporated with a scaling factor $\gamma$. Finally, the advantages in the global batch are normalized and used for GRPO training:
\vspace{-0.2\baselineskip}
\begin{equation}
A_\text{raw} = R_\text{score} - \mathrm{mean}_\text{group}(R_\text{score}) + \gamma \cdot R_\text{format},
\label{eq:adv_cal}
\end{equation}
\vspace{-0.5\baselineskip}
\begin{equation}
A=\frac{A_\text{raw}-\mathrm{mean}_\text{batch}(A_\text{raw})}{\mathrm{std}_\text{batch}(A_\text{raw})}.
\label{eq:adv_norm}
\end{equation}

\subsubsection{Layout Reward Model}
\label{subsec:rm_train}
Training a robust reward model for text layout is uniquely challenging. Unlike tasks with binary correctness, layout quality is a holistic and fine-grained judgment of aesthetics, readability, and coherence. A high-quality preference dataset is crucial for training a model that can guide a stable RL without falling into reward hacking. However, no existing datasets or methodologies are specifically designed for layout generation reward modeling. 
To address this gap, we train a specialized reward model $r_{\theta}$ that takes triplets $(B, T, I)$ as input and outputs a scalar score $R_{\text{layout}}$, where $B$ denotes the background image, $T$ denotes the target text, and $I$ denotes the rendered layout image.

\textbf{Dataset:} To create the necessary preference data, we introduce a novel hierarchical data construction method that creates fine-grained quality distinctions across multiple layout quality levels.

To equip the reward model with comprehensive and robust evaluation capabilities, we construct four distinct quality levels to capture fine-grained layout performance:
\begin{itemize}
    \item \textbf{Level-I:} High-quality ground-truth layouts serving as gold standards for design excellence.
    \item \textbf{Level-II:} Layouts generated by Qwen2.5-VL-7B after fine-tuning on 200K samples exhibit reasonable quality with minor imperfections.
    \item \textbf{Level-III:} Layouts from the Level-II model subjected to systematic spatial perturbations applied to layout elements, including random positional offsets that moderately compromise design coherence.
    \item \textbf{Level-IV:} Severely degraded layouts stem from the Level-II model under aggressive perturbations, including extensive positional displacement, random font size variations, selective text element deletion, image reference removal, and arbitrary SVG scaling transformations.
\end{itemize}

This hierarchical construction enables comprehensive preference learning through systematic pairwise comparisons. For each layout generation prompt, we create layouts at all four quality levels, then form all possible pairwise comparisons between different levels. This yields 6 preference pairs per problem, establishing clear quality orderings that capture nuanced distinctions essential for effective reward model training. This strategy forces the reward model to move beyond simple binary (good/bad) judgments and learn the subtle distinctions that separate excellent layouts from merely acceptable ones. The resulting dataset provides a comprehensive and reliable basis for training a highly discerning layout reward model $r_{\theta}$.

\textbf{Training:} The method proposed in~\cite{ouyang2022training} is adopted for training the reward model. Specifically, the reward model is initialized with Qwen2.5-VL-3B. To adapt it for preference learning, the final layer is replaced with a linear layer yielding a scalar output. Subsequently, the reward model is trained via the negative log-likelihood loss function:
\begin{equation}
  \resizebox{\linewidth}{!}{$%
\mathcal{L}_{\mathrm{RM}}(\theta) = -\mathbb{E}_{(B, T, I^{+}, I^{-}) \sim \mathcal{D}} \left[ \log\sigma\left(r_{\theta}\left(B, T, I^{+}\right) - r_{\theta}\left(B, T, I^{-}\right)\right) \right]$,}
\end{equation}
where $I^{+}$ and $I^{-}$ denote the better-quality and lower-quality images, respectively, in a pairwise comparison.

To provide stable reward values for the RL process, the raw output of $r_\theta$ is normalized before being used as the final layout reward $R_\text{layout}$. Following the practice in \cite{xu2023imagereward}, we standardize $r_\theta$ scores using the mean and standard deviation of the test set distribution. This procedure ensures that the reward signal maintains a consistent scale throughout training, which is crucial for effective policy optimization. The final reward is calculated as:
\begin{equation}
R_\text{layout} = \frac{r_\theta - \mathrm{mean}_\text{test}(r_\theta)}{\mathrm{std}_\text{test}(r_\theta)}.
\label{rm_norm}
\end{equation}

\section{Experiment}
\label{sec:experiment}

\subsection{Experimental Setup} 

\noindent \textbf{Evaluation Datasets}
For evaluation, we randomly sample 1K examples from our TextLayout dataset as the primary test set, ensuring no overlap with training data. We also conduct additional experiments on the Crello~\cite{yamaguchi2021canvasvae} and DESIGNERINTENTION~\cite{jia2023cole} benchmarks, preprocessed into background-text pairs.

\noindent \textbf{Evaluation metrics}
We adopt three groups of evaluation metrics. Text accuracy is measured using character-level F-measure based on OCR recognition. For graphic metrics, we adopt $R_{ali}$, $R_{ove}$, and $R_{com}$~\cite{zhou2022composition, qu2025igd}, which capture text position alignment, text-text overlap, and pixel gradient smoothness within text regions. Additionally, we employ GPT-4o as a judge~\cite{haraguchi2024can} to evaluate four dimensions: Text Accuracy, Text-Background Harmony, Text Presentation Quality, and Meaning Expression Adaptability. Refer to the supplementary materials for details.

\noindent \textbf{Existing Methods}
We compare against three categories of existing methods: (1) Advanced MLLMs: GPT-4o~\cite{hurst2024gpt}, Claude 3.7~\cite{anthropic2025claude}, Doubao-Seed-1.6\cite{doubao1.6}, and Qwen2.5-VL-72B~\cite{qwen2.5-vl}; (2) Image editing models: GPT-4o(edit)~\cite{hurst2024gpt}, Qwen-Image-Edit~\cite{wu2025qwenimagetechnicalreport}, and FLUX-Kontext~\cite{labs2025flux}; (3) Specialized layout generation: the open-source domain-specific model OpenCOLE~\cite{inoue2024opencole} and IGD~\cite{qu2025igd}.

\begin{table*}[htbp]
\centering
\caption{The Graphic quality metrics and OCR metrics on the TextLayout, Crello and DESIGNERINTENTION test set, where VFLM-step1 and VFLM respectively represent the metrics of our results in the first output and the final result output after iterative reflection.}  
\label{tab:exp_main_all}   
\resizebox{\textwidth}{!}
{
\begin{tabular}{cc|c|ccc|c|ccccc}
\hline
\multirow{2}{*}{Method}        & \multirow{2}{*}{Model} & \multicolumn{1}{c|}{OCR} & \multicolumn{3}{c|}{Graphic}                                        &                       & \multicolumn{5}{c}{GPT-4o}                                                                                      \\
                               &                        & Char-F $\uparrow$       & $R_{ali} \downarrow$ & $R_{ove} \downarrow$ & $R_{com} \downarrow$  & RM Score $\uparrow$   & Text                 & Harmony              & Quality              & Meaning              & Overall              \\ \hline
\rowcolor[HTML]{E7F8FE} \multicolumn{12}{l}{TextLayout}                                                                                                                                                                                                                                            \\ \hline
\multirow{5}{*}{MLLM}          & GPT-4o                 & 0.8258                  & 0.0046               & {\ul 0.0033}         & 18.8443               & 0.3561                & 8.5165               & 8.0341               & 7.2826               & 7.6553               & 7.8721               \\
                               & Claude3.7              & 0.8672                  & 0.0053               & 0.0383               & 16.4401               & 0.5295                & 8.8058               & 8.4026               & 7.7691               & 8.2651               & 8.3106               \\
                               & Doubao-Seed-1.6        & 0.8484                  & 0.0058               & 0.0216               & 18.5823               & 0.4063                & 8.5663               & 8.2304               & 7.4463               & 7.8917               & 8.0337               \\
                               & Qwen2.5-VL-72B             & 0.7178                  & {\ul 0.0031}         & 0.0358               & 19.9399               & 0.1989                & 7.8633               & 7.6094               & 6.5402               & 6.5582               & 7.1428               \\
                               & Qwen2.5-VL-7B             & 0.6195                  & \textbf{0.0029}      & 0.0229               & 25.7715               & 0.1166                & 7.5638               & 6.4431               & 5.4827               & 5.3861               & 6.5489               \\ \hline
\multirow{3}{*}{Image Edit}    & GPT-4o                 & 0.7198                  & -                    & -                    & -                     & 0.3371                & 7.4809               & \textbf{8.9298}      & \textbf{8.1672}      & 8.1919               & 8.1924               \\
                               & Qwen-Image-Edit        & 0.7256                  & -                    & -                    & -                     & 0.3062                & 5.4086               & 7.7783               & 6.1886               & 6.0040               & 6.3449               \\
                               & FLUX Kontext           & 0.1322                  & -                    & -                    & -                     & -0.5853               & 1.6473               & 6.8098               & 3.6345               & 2.2142               & 3.5765               \\ \hline
\multirow{2}{*}{Layout}        & OpenCOLE               & 0.2147                  & 1.2029               & 0.0316               & 25.1150               & 0.0397                & 2.6596               & 6.3873               & 3.7020               & 2.8896               & 3.9096               \\
                               & IGD                    & 0.8481                  & 0.0057               & 0.0114               & {\ul 14.8597}         & 0.3646                & 8.4324               & 7.8125               & 7.3055               & 7.6894               & 7.8100               \\ \hline
                               & VFLM-step1             & {\ul 0.9071}            & 0.0035               & 0.0059               & 15.4583               & {\ul 0.5415}          & {\ul 8.8880}         & 8.3591               & 7.7255               & {\ul 8.2896}         & {\ul 8.3155}         \\
\rowcolor[HTML]{F0F0F0}        & VFLM & \textbf{0.9376}         & 0.0039               & \textbf{0.0009}      & \textbf{11.8678}      & \textbf{0.6018}       & \textbf{9.0447}      & {\ul 8.7492}         & {\ul 7.9679}         & \textbf{8.5969}      & \textbf{8.5897}      \\
\rowcolor[HTML]{D8DAD9} \multirow{-3}{*}{Ours}           & $\Delta$ (vs step1)    & +0.0305                 & -0.0004              & +0.005               & +3.5905               & +0.0603               & +0.1567              & +0.3901              & +0.2424              & +0.3073              & +0.2742              \\ \hline
\rowcolor[HTML]{E7F8FE} \multicolumn{12}{l}{Crello}                                                                                                                                                                                                                                                \\ \hline
\multirow{5}{*}{MLLM}          & GPT-4o                 & {\ul 0.9101}            & \textbf{0.0017}      & {\ul 0.0036}         & 19.9216               & 0.4427                & {\ul 8.9612}         & 7.9949               & 7.5237               & 7.9618               & 8.1104               \\
                               & Claude3.7              & 0.8610                  & 0.0058               & 0.0205               & 21.0798               & \textbf{0.5880}       & \textbf{8.9645}      & 8.3805               & {\ul 7.8999}         & {\ul 8.4507}         & {\ul 8.4239}         \\
                               & Doubao-Seed-1.6        & 0.8758                  & 0.0051               & 0.0192               & 21.5918               & 0.4876                & 8.8927               & 8.2174               & 7.6658               & 8.2569               & 8.2582               \\
                               & Qwen2.5-VL-72B             & 0.8115                  & 0.0044               & 0.0619               & 24.6138               & 0.2359                & 7.7332               & 7.4732               & 6.5589               & 6.5405               & 7.0764               \\
                               & Qwen2.5-VL-7B             & 0.7620                  & 0.0028               & 0.0256               & 30.9031               & 0.2116                & 7.6289               & 6.2840               & 5.5726               & 5.5813               & 6.2667               \\ \hline
\multirow{3}{*}{Image Edit}    & GPT-4o                 & 0.8774                  & -                    & -                    & -                     & {\ul 0.5849}          & 8.8438               & \textbf{9.0127}      & \textbf{8.8051}      & \textbf{9.0069}      & \textbf{8.9171}      \\
                               & Qwen-Image-Edit        & 0.8326                  & -                    & -                    & -                     & 0.3671                & 5.7509               & 7.2808               & 6.0635               & 5.9903               & 6.2714               \\
                               & FLUX Kontext           & 0.5558                  & -                    & -                    & -                     & -0.0834               & 3.2447               & 6.6992               & 4.5005               & 4.1499               & 4.6486               \\ \hline
\multirow{2}{*}{Layout}        & OpenCOLE               & 0.6411                  & 1.1902               & 0.0429               & 31.5831               & 0.3345                & 6.9777               & 7.3864               & 6.4068               & 6.5319               & 6.8257               \\
                               & IGD                    & 0.8853                  & 0.0128               & 0.0182               & 20.2092               & 0.3959                & 8.3001               & 7.5402               & 6.9126               & 7.4004               & 7.5383               \\ \hline
                               & VFLM-step1             & 0.8774                  & 0.0046               & 0.0061               & {\ul 19.5917}         & 0.4392                & 8.2334               & 7.9055               & 7.2046               & 7.6543               & 7.7494               \\
\rowcolor[HTML]{F0F0F0}        & VFLM & \textbf{0.9256}         & {\ul 0.0025}         & \textbf{0.0022}      & \textbf{14.8063}      & 0.5548                & 8.8260               & {\ul 8.5957}         & 7.7267               & 8.2964               & 8.3562               \\
\rowcolor[HTML]{D8DAD9} \multirow{-3}{*}{Ours}           & $\Delta$ (vs step1)    & +0.0482                 & +0.0021              & +0.0039              & +4.7854               & +0.1156               & +0.5926              & +0.6902              & +0.5221              & +0.6421              & +0.6068              \\ \hline
\rowcolor[HTML]{E7F8FE} \multicolumn{12}{l}{DESIGNERINTENTION}                                                                                                                                                                                                                                     \\ \hline
\multirow{5}{*}{MLLM}          & GPT-4o                 & {\ul 0.9563}            & {\ul 0.0015}         & 0.0140               & 18.7249               & 0.5072                & 9.5180               & 8.2725               & 8.0200               & 8.4820               & 8.5731               \\
                               & Claude3.7              & 0.8753                  & 0.0070               & 0.0601               & 15.0869               & \textbf{0.6465}       & 9.5815               & {\ul 8.5473}         & {\ul 8.2052}         & {\ul 8.8229}         & {\ul 8.7892}         \\
                               & Doubao-Seed-1.6        & 0.8917                  & 0.0046               & 0.0222               & 15.7470               & 0.5103                & 9.1440               & 8.0020               & 7.7400               & 8.2960               & 8.2955               \\
                               & Qwen2.5-VL-72B             & 0.5475                  & 0.0069               & 0.1649               & 17.8489               & 0.3104                & 4.1891               & 7.7565               & 6.8813               & 4.4970               & 5.8310               \\
                               & Qwen2.5-VL-7B             & 0.6694                  & \textbf{0.0011}      & 0.0689               & 19.1119               & 0.2018                & 7.3688               & 6.1285               & 5.6576               & 5.4498               & 6.1512               \\ \hline
\multirow{3}{*}{Image Edit}    & GPT-4o                 & 0.8832                  & -                    & -                    & -                     & {\ul 0.6278}          & {\ul 9.5231}         & \textbf{8.9287}      & \textbf{9.0126}      & \textbf{9.2558}      & \textbf{9.1800}      \\
                               & Qwen-Image-Edit        & 0.8518                  & -                    & -                    & -                     & 0.5995                & 7.1626               & 7.7435               & 7.2625               & 7.1403               & 7.3272               \\
                               & FLUX Kontext           & 0.5412                  & -                    & -                    & -                     & 0.4426                & 4.3908               & 7.2806               & 5.4208               & 5.1503               & 5.5606               \\ \hline
\multirow{2}{*}{Layout}        & OpenCOLE               & 0.7308                  & 0.6848               & 0.0408               & 20.2853               & 0.4684                & 8.4738               & 7.3488               & 6.9516               & 7.4435               & 7.5544               \\
                               & IGD                    & 0.9283                  & 0.0115               & 0.0065               & 14.5905               & 0.5349                & 8.8112               & 7.7505               & 7.4449               & 7.8677               & 7.9686               \\ \hline
                               & VFLM-step1  & 0.9415                  & 0.0033               & {\ul 0.0023}         & {\ul 14.1230}         & 0.5285                & 9.4020               & 8.1924               & 7.8640               & 8.3260               & 8.4461               \\
\rowcolor[HTML]{F0F0F0}        & VFLM & \textbf{0.9663}         & 0.0024               & \textbf{0.0008}      & \textbf{12.1167}      & 0.5688                & \textbf{9.5569}      & 8.4511               & 7.9301               & 8.5130               & 8.6128               \\
\rowcolor[HTML]{D8DAD9} \multirow{-3}{*}{Ours}          & $\Delta$ (vs step1)    & +0.0248                 & +0.0009              & +0.0015              & +2.0063               & +0.0403               & +0.1549              & +0.2587              & +0.0661              & +0.1870              & +0.1667              \\ \hline
\end{tabular}
}
\vspace*{-\baselineskip}
\end{table*}
\begin{figure*}[ht]
    \centering
    \includegraphics[width=0.99\linewidth]{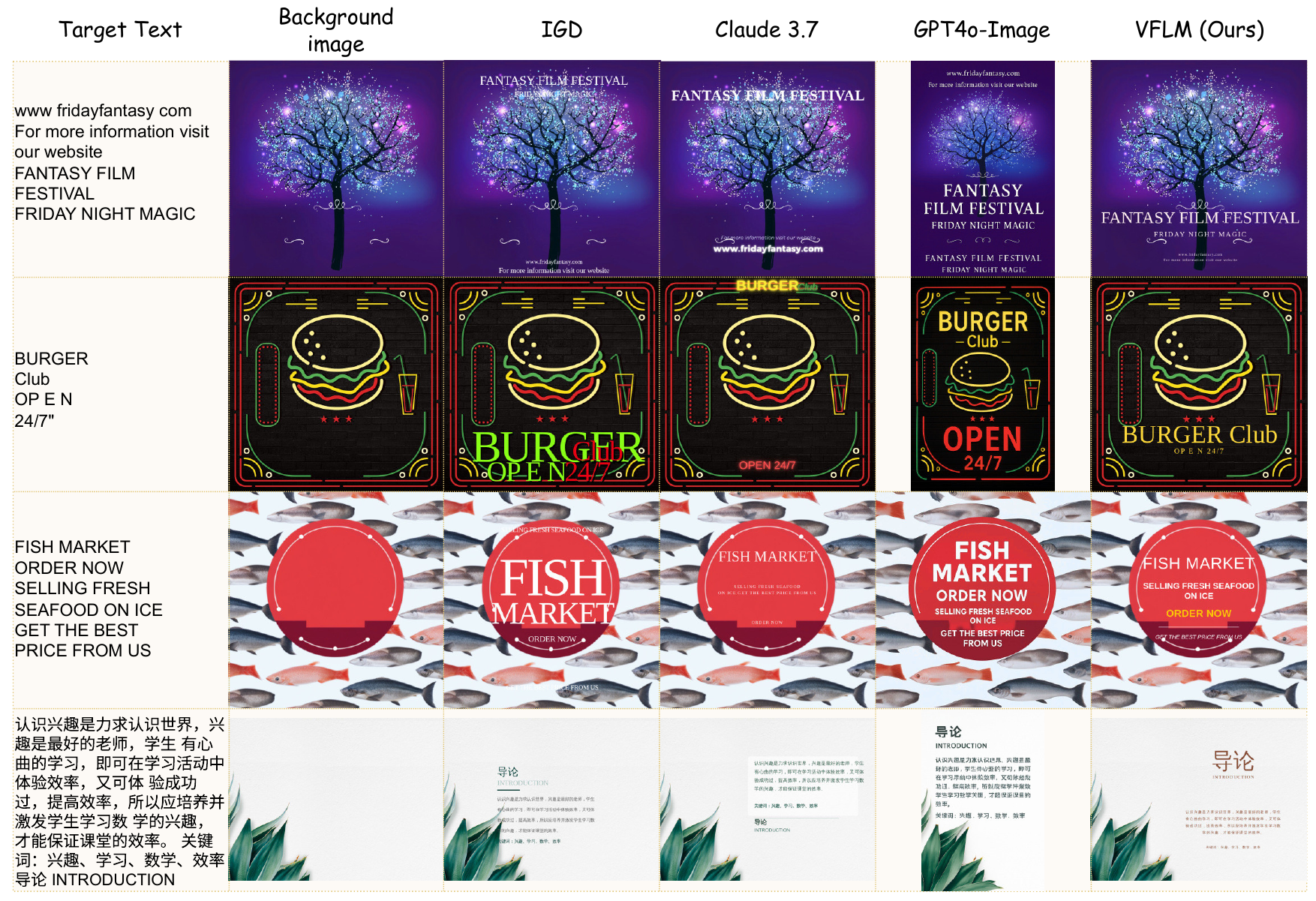}
    \caption{In comparison with existing methods, we selected one model from each category of methods as a representative. For a more comprehensive comparison, please refer to the supplementary materials.} 
    \label{fig:compare}
    \vspace*{-\baselineskip}
\end{figure*}
\subsection{Comparison with Existing Methods}

As shown in \cref{tab:exp_main_all}, VFLM demonstrates a clear quantitative advantage. On the TextLayout dataset, VFLM's initial generation (VFLM-step1) already achieves a higher OCR F1-score (0.9071) than the strongest MLLM competitor, Claude 3.7 (0.8672), and the specialized layout model IGD (0.8481). The final optimized output (VFLM) widens this gap further to 0.9376. This superiority extends across key graphic metrics: VFLM excels at minimizing element overlap ($R_\text{ove}$) and optimizing composition ($R_\text{com}$), achieving the highest RM scores and overall GPT-4o evaluation. This strong performance generalizes robustly: VFLM consistently achieves state-of-the-art results on the Crello and DESIGNERINTENTION datasets, establishing a new SOTA on nearly all OCR and Graphic metrics.

\cref{fig:compare} provides a qualitative comparison of VFLM against existing methods. Models like IGD and Claude 3.7, which generate layouts directly, inevitably suffer from visual artifacts such as text overlap and color conflicts. Meanwhile, image editing models like GPT4o-Image fundamentally conflict with our task, as they inevitably alter the background image. These editing models also struggle to guarantee quality in scenarios with dense text or Chinese characters. VFLM, in contrast, successfully integrates the element-wise consistency of code generation while further leveraging visual feedback to promptly discover and correct visual problems. This ensures a high-quality final output, overcoming the limitations of both competing approaches.

\subsection{Layout Reward Model Evaluation}
\label{sec:rm_eval}

The trained reward model achieves a high pairwise prediction accuracy of 97.4\% on the preference data test set.
To further validate our methodology, we conducted two key analyses presented in \cref{tab:rm_eval}. First, we verify the integrity of our four-level data hierarchy using objective metrics. As shown, the external Graphic and OCR metrics exhibit a clear monotonic degradation from Level-I to Level-IV. Among them, $R_\text{ove}$ and $R_\text{com}$ have slight improvement in Leval-III and Leval-IV because many texts in these two types of data have been lost in the layout, so text overlap and regional gradients will naturally decrease. This result confirms that our data construction process successfully creates a well-defined and reliable quality gradient.
\begin{table}[htbp]
\centering
\caption{Metrics and RM score across four quality-level datasets.}  
\label{tab:rm_eval}   
\resizebox{\linewidth}{!}
{
\begin{tabular}{c|c|ccc|c}
\hline
\multirow{2}{*}{level} & OCR               & \multicolumn{3}{c|}{Graphic}                                       & \multirow{2}{*}{RM Score $\uparrow$} \\
                       & Char-F $\uparrow$ & $R_{ali} \downarrow$ & $R_{ove} \downarrow$ & $R_{com} \downarrow$ &                                      \\ \hline
Level-I                & 0.9474            & 0.0089               & 0.0038               & 6.6795               & 1.0594                               \\
Level-II               & 0.9049            & 0.0112               & 0.0134               & 12.2626              & 0.6343                               \\
Level-III              & 0.5657            & 0.0354               & 0.0444               & 21.2628              & -0.2345                              \\
Level-IV               & 0.5324            & 0.0536               & 0.0375               & 17.5270              & -1.3457                              \\ \hline
\end{tabular}
}
\vspace*{-\baselineskip}
\end{table}

Second, we evaluated whether our trained reward model internalizes this quality structure. The final column of the table reports the average RM Score, i.e., $R_\text{layout}$, which aligns well with the established hierarchy, decreasing consistently from a high for Level-I to a low for Level-IV. This strong discriminative performance across distinct quality levels confirms that our model has learned a nuanced understanding of layout quality, enabling it to provide a reliable and effective supervision signal for reinforcement learning.

\subsection{Ablation Study}
In the ablation study, we conducted comparative experiments using TextLayout test set. 
To comprehensively compare the effects of our visual feedback method, we compared it against several training approaches: (1) \textbf{Cold-Start Model}: The cold start model mainly ensures the format of the iterative output and does not significantly improve the layout quality; 
(2) \textbf{Single-Round RL}: We trained the cold-start model using RL but restricted generation to only one step, enabling fair comparison between single-round generation and iterative visual reflection; (3) \textbf{RL from Pre-trained Models}: Direct RL training from the pre-trained Qwen2.5-VL-7B model without SFT initialization; (4) \textbf{Direct Output}: SFT+RL training was performed using the same source data as our visual feedback method for direct SVG code generation; (5) \textbf{Direct Output SFT}: For fair comparison with our 40K-sample SFT+RL approach, we trained a direct SFT model on 40K samples. Since these models only require a single round of generation training, we use the standard SFT loss and GRPO Advantage calculation method. For more ablation experiments on reward functions and advantage algorithms, as well as training details, please refer to the supplementary materials.
\begin{table}[htbp]
\centering
\caption{Ablation experiments on Graphic quality metrics and OCR metrics on TextLayout test set.}  
\label{tab:ablation}   
\resizebox{\linewidth}{!}{
\begin{tabular}{c|c|ccc|c}
\hline
\multirow{2}{*}{Model}   & OCR               & \multicolumn{3}{c|}{Graphic}                                       & \multicolumn{1}{l}{\multirow{2}{*}{RM Score $\uparrow$}} \\
                         & Char-F $\uparrow$ & $R_{ali} \downarrow$ & $R_{ove} \downarrow$ & $R_{com} \downarrow$ & \multicolumn{1}{l}{}                                     \\ \hline
VFLM-step1    & 0.9071            & 0.0035               & 0.0059               & 15.4583              & {\ul 0.5415}                                             \\
VFLM   & \textbf{0.9376}   & 0.0039               & \textbf{0.0009}      & \textbf{11.8678}     & \textbf{0.6018}                                          \\ \hline
Cold-Start-step1         & 0.7913            & 0.0078               & 0.0297               & 19.0560              & 0.2572                                                   \\
Cold-Start        & 0.7980            & 0.0081               & 0.0298               & 19.0577              & 0.2608                                                   \\ \hline
Single-Round RL & 0.8792            & {\ul 0.0024}         & 0.0053               & 18.9428              & 0.4063                                                   \\
RL from Pretrained       & 0.8154            & \textbf{0.0003}      & 6.7109               & 22.0273              & 0.2971                                                   \\
Direct Output            & {\ul 0.9237}      & 0.0027               & {\ul 0.0021}         & 17.0654              & 0.4964                                                   \\
Direct Output SFT        & 0.8551            & 0.0040               & 0.0153               & {\ul 12.9459}        & 0.5332                                                   \\ \hline
\end{tabular}
}
\end{table}

The results of our ablation study, presented in \cref{tab:ablation}, demonstrate the clear superiority of our Visual Feedback method. Our VFLM achieves the best performance across the majority of metrics, substantially outperforming all RL baselines. More notably, the quality of our initial output (VFLM-step1) is already not inferior to any other competing methods. For instance, its RM score of 0.5415 even surpasses the second-best performing Direct Output SFT (0.5332). Subsequent iterative steps further widened this performance gap, increasing the RM score to 0.6018, and achieving top-notch results in both OCR and image quality. Our success highlights that our visual feedback framework is a more effective solution for layout generation, as it can establish a higher quality benchmark from the very first step and then optimize it to the state-of-the-art level.

\begin{figure*}[!ht]
  \centering
  \begin{minipage}{0.33\textwidth}
    \centering
    \includegraphics[width=\linewidth]{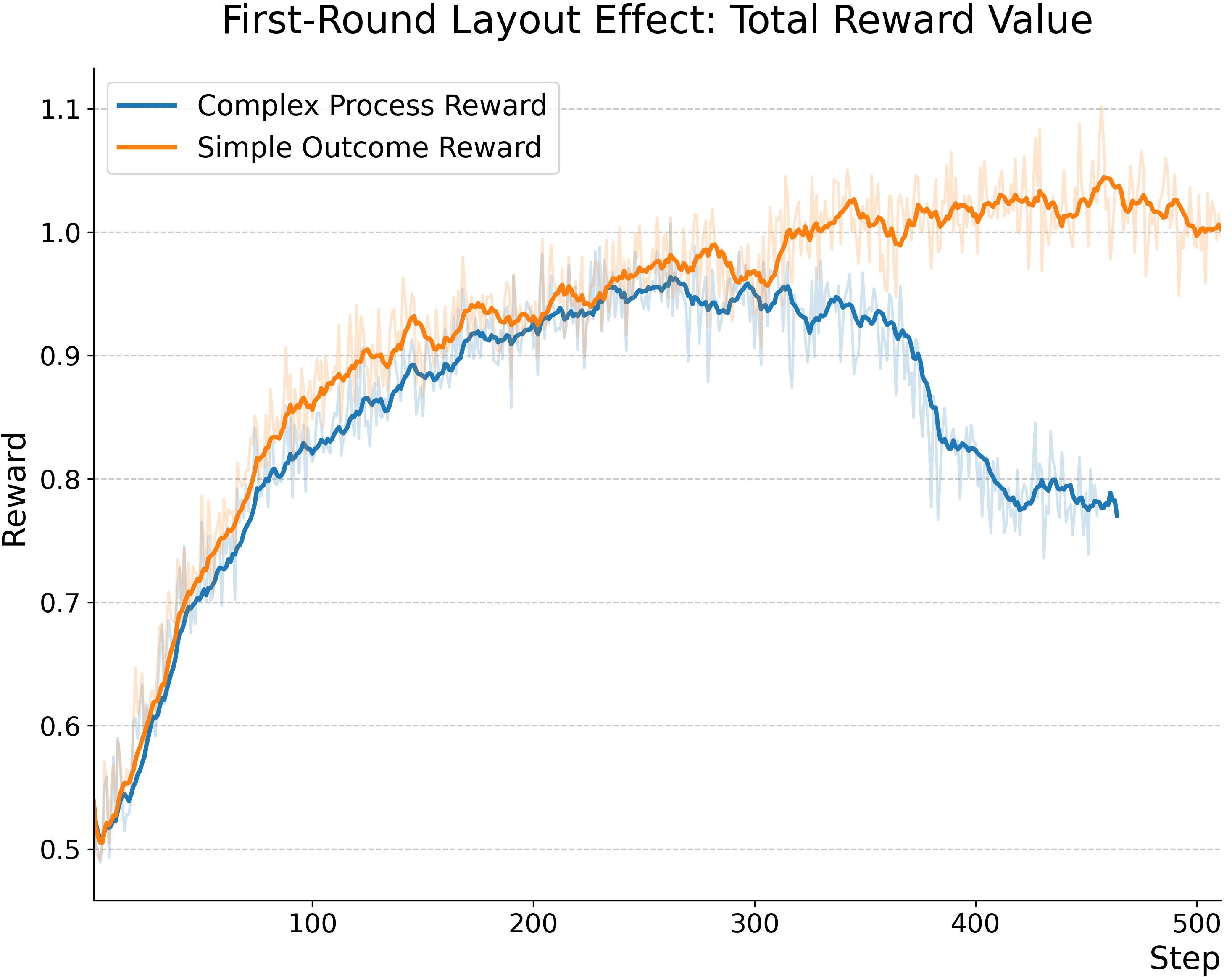}
  \end{minipage}
  \begin{minipage}{0.33\textwidth}
    \centering
    \includegraphics[width=\linewidth]{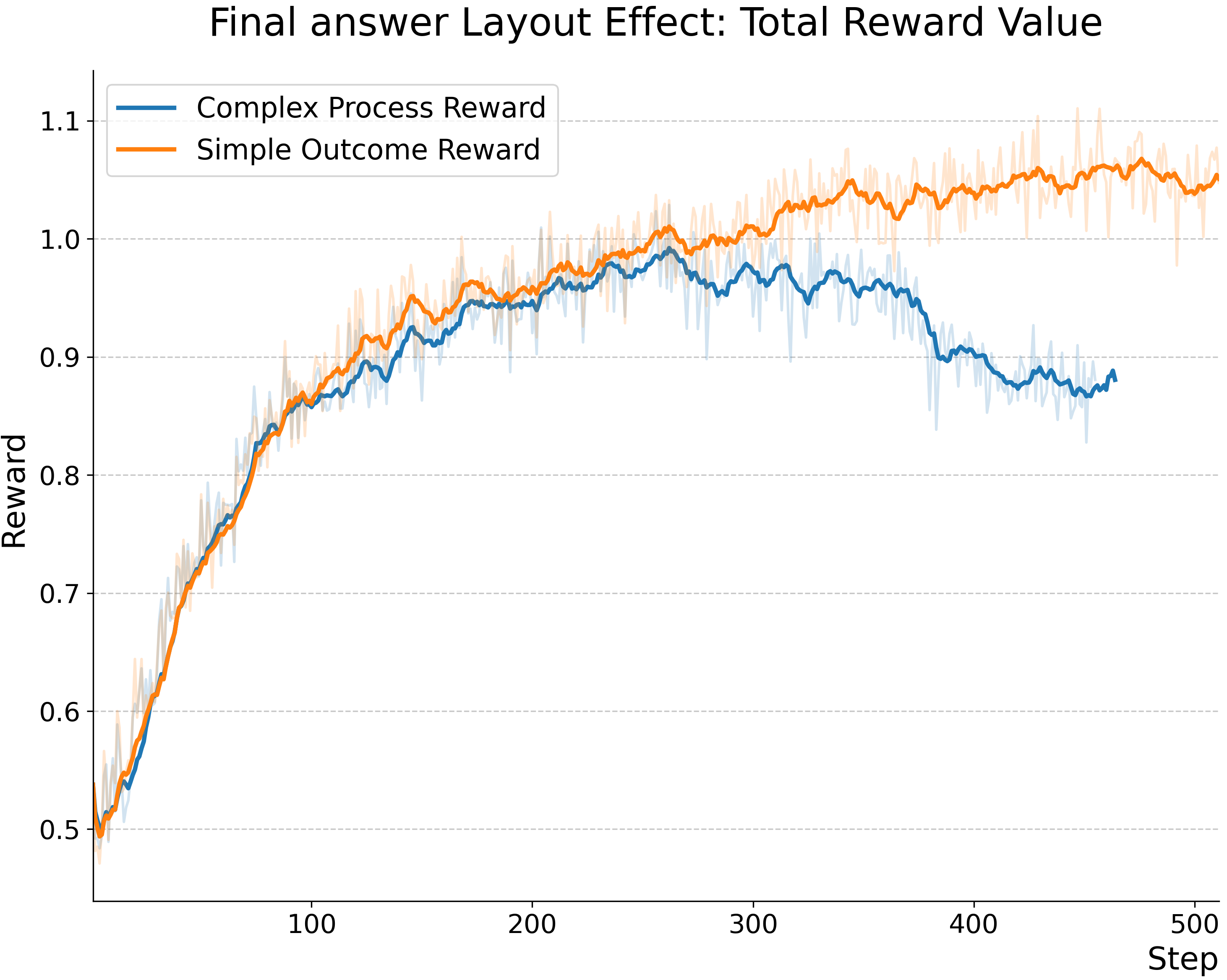}
  \end{minipage}
  \begin{minipage}{0.33\textwidth}
    \centering
    \includegraphics[width=\linewidth]{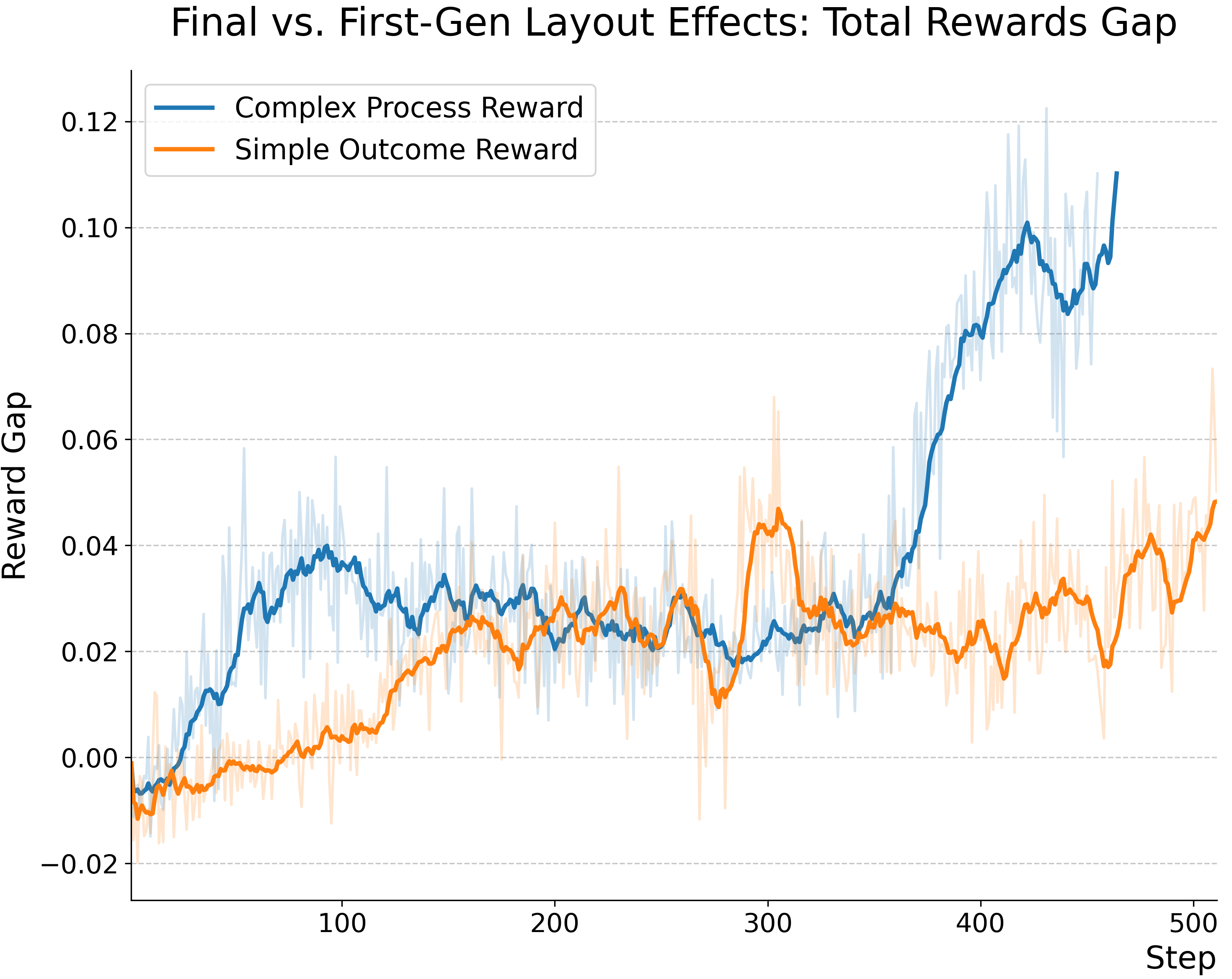}
  \end{minipage}
  \caption{Comparison of training processes: simple outcom rewards vs. complex process rewards.}
  \label{fig:rl_dynamic}
  \vspace*{-\baselineskip}
\end{figure*}

\subsection{Discussion: Effectiveness of Simple Rewards}
\label{sec:discussion}

\textbf{Can simple outcome-based rewards effectively stimulate self-improvement capabilities in MLLMs?} Our empirical investigation provides compelling evidence that they can, and even outperform more sophisticated alternatives.

We designed a sophisticated process-oriented reward function (\cref{eq:Aq_ot}–\cref{eq:R_length}). This complex reward scheme incorporates three distinct optimization objectives: (1) first-round quality maximization using group-wise mean baselines, (2) iterative improvement encouragement via maximum-quality baselines from prior rounds, and (3) strategic termination control via reward and length penalties to prevent reward hacking. Subsequent advantages will be normalized through \cref{eq:adv_norm}.
\begin{equation}
A_{q, o_{t_i}} = 
\begin{cases} 
R_{q, o_{t_1}} - \text{mean}_\text{group}(R_{q, o_{t_1}}) & \text{if } i = 1, \\
2 \cdot \left( R_{\text{answer}} + R_{\text{length}} \right) & \text{if } i = \text{last}, \\
2 \cdot \left( R_{q, o_{t_i}} - \max(R_{q, o_{t_{\leq i}}}) \right) & \text{else} ,
\end{cases}
\label{eq:Aq_ot}
\end{equation}
where $o_t$ represents a complete generation path, and $o_{t_i}$ represents the response of the $i$-th round in this complete path. The terminal reward ($i=\text{last}$) components are defined as:
\begin{equation}
R_{\text{answer}} = 
\begin{cases} 
0.7 \qquad \quad \text{if } R_{q, o_{t_\text{last}}} \geq \max(R_{q, o_{t_{\leq \text{last}}}}),  \\
R_{q, o_{t_{\text{last}}}} - \max(R_{q, o_{t_{\leq \text{last}}}}) \qquad \quad \text{ else },
\end{cases}
\label{eq:R_answer}
\end{equation}
\begin{equation}
\begin{split}
R_{\text{length}} = -2 \cdot \bigg[ &\left( R_{q, o_{t_\text{last}}} - \text{max}_\text{group}(R_{q, o_{t_\text{last}}}) \right) \\
&\cdot \text{max}(0, 4-\mathrm{tool\_call\_count}) \bigg].
\end{split}
\label{eq:R_length}
\end{equation}

\cref{fig:rl_dynamic} illustrates the training dynamics of the two reward schemes. Before 250 steps, both algorithms improved stably: their final answer scores (middle subfigure) showed nearly identical trends, and first-round generation performance was comparable, the outcome-only reward algorithm even slightly outperform the complex process RL one. After 250 steps, the process RL converged and even suffered performance degradation, whereas the outcome-only reward continued to improve steadily until training concluded. The reward gap (right subfigure) reveals that process RL quickly widened this gap but plateaued later, which is likely due to restricted first-round learning in early training. In contrast, the outcome-only reward gradually mastered progressive iterative refinement. \cref{tab:agentic_rl} quantifies these observations through comprehensive evaluation metrics. The simple outcome-based reward demonstrates superior effectiveness in stimulating self-improvement capabilities across all evaluation dimensions.
\begin{table}[htbp]
\centering
\caption{Comparison of simple outcome rewards and complex process rewards on our test set.}  
\label{tab:agentic_rl}   
\resizebox{\linewidth}{!}
{
\begin{tabular}{c|c|ccc|c}
\hline
                                                                 & \multicolumn{1}{c|}{OCR}                                  & \multicolumn{3}{c|}{Graphic}                                                              &                                       \\
\multirow{-2}{*}{level}                                          & Char-F $\uparrow$ & $R_{ali} \downarrow$ & $R_{ove} \downarrow$ & {\color[HTML]{000000} $R_{com} \downarrow$} & \multirow{-2}{*}{RM Score $\uparrow$} \\ \hline
\rowcolor[HTML]{FFFFFF} 
Outcome RL-step1                                             & 0.9071            & \textbf{0.0035}               & 0.0059               & {\ul 15.4583}                                     & {\ul 0.5415}                                \\
\rowcolor[HTML]{FFFFFF} 
Outcome RL                                            & \textbf{0.9376}            & {\ul 0.0039}               & \textbf{0.0009}               & \textbf{11.8678}                                     & \textbf{0.6018}                                \\
\rowcolor[HTML]{D8DAD9} 
\multicolumn{1}{r|}{\cellcolor[HTML]{D8DAD9}$\Delta$ (vs step1)} & +0.0305            & -0.0004              & +0.005                & +3.5905                                      & +0.0603                                \\ \hline
\rowcolor[HTML]{FFFFFF} 
Process RL-step1                                                 & 0.9032            & 0.0053               & 0.0030               & 16.6528                                     & 0.4936                                \\
\rowcolor[HTML]{FFFFFF} 
Process RL                                                & {\ul 0.9239}            & 0.0052               & {\ul 0.0027}               & 16.4220                                     & 0.5241                                \\
\rowcolor[HTML]{D8DAD9} 
\multicolumn{1}{r|}{\cellcolor[HTML]{D8DAD9}$\Delta$ (vs step1)} & +0.0207            & +0.0001               & +0.0003               & +0.2308                                      & +0.0305                                \\ \hline
\end{tabular}
}
\end{table}
These findings reveal a fundamental insight: under effective visual feedback mechanisms, simple outcome-based rewards can successfully harness the inherent visual understanding capabilities of MLLMs to elicit robust self-improvement behaviors, while complex process-oriented rewards may actually inhibit optimal performance, leading to the brittle local optima and ``reward hacking'' (policy collapse) observed in \cref{fig:rl_dynamic}. This counterintuitive result suggests that the powerful internal representations and reasoning capabilities of modern MLLMs, when properly guided by clear outcome objectives and visual feedback, can autonomously develop sophisticated iterative refinement strategies without explicit process supervision.
\section{Conclusion}
\label{sec:conclusion}
This paper introduces VFLM, a self-improving framework for text layout design, bridging the critical ``visual perception gap'' of existing code-only methods. We implement a ``generation, rendering, reflection, refinement'' loop, which is trained via RL that leverages a novel, specialized layout reward model. A key finding is that rewarding only the final outcome, rather than complex intermediate steps, can more effectively stimulate the model's self-improvement capabilities. Extensive experiments validate that VFLM consistently outperforms code-only methods, establishing visual feedback as essential for automated design and MLLM-based design agents.

\section*{Acknowledgments}
This work is supported by the the National Nature Science Foundation of China (62425114, 62121002, U23B2028), and the Fundamental and Interdisciplinary Disciplines Breakthrough Plan of the Ministry of Education of China (JYB2025XDXM103).

{
    \small
    \bibliographystyle{ieeenat_fullname}
    \bibliography{main}
}

\clearpage
\setcounter{page}{1}
\maketitlesupplementary
\setcounter{section}{0}     

\section{Dataset Details}
\label{sec:dataset_detail}
\begin{figure}[h]
    \centering
    \includegraphics[width=1\linewidth]{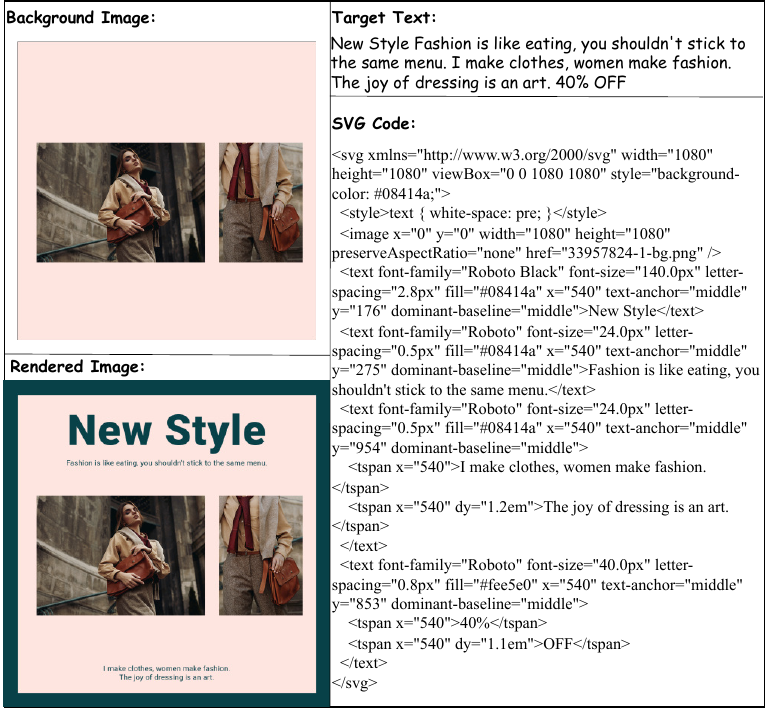}
    \caption{Data example, consisting of a background image, target text, SVG layout code, and the corresponding rendered image.}
    \label{fig:data_example}
\end{figure}

We collected approximately 200K samples, including free and paid data from the internet. Each sample contains a background image, target text, well-formatted SVG code, and the corresponding rendered image. Each sample has an average of 9.8 text boxes, with an average text length of 84.7. \cref{fig:data_example} is a data example. Based on these data, our training and testing data were constructed, including the dataset for the Cold-Start SFT phase, the queries for the reinforcement learning phase, the training and testing datasets for the reward model, and finally, the dataset used for evaluation. 

\subsection{Cold-Start Dataset Details}
We use the method described in the main text and set the maximum number of modifications to 3 during the data generation process. Since each data trajectory includes an initial generation and a final "satisfied" output, the total number of turns is the number of modifications plus two. This process yielded a total of 8K trajectories, comprising 2,359 two-turn samples (0 modifications), 1,266 three-turn samples (1 modification), 2,030 four-turn samples (2 modifications), and 2,537 five-turn samples (3 modifications). Total data volume is 8K. 

During cold-start data construction, the prompts used to guide Doubao-Seed-1.6 are prompt~\ref{prompt:Initial_Reasoning} and prompt~\ref{prompt:Multi_round_reflection}, where prompt~\ref{prompt:Initial_Reasoning} is utilized for Initial Reasoning synthesis and prompt~\ref{prompt:Multi_round_reflection} for Multi-round reflection synthesis.

\subsection{Layout Reward Model Dataset}
Herein, a qualitative comparison of data across the four quality levels (Level-I, Level-II, Level-III, Level-IV) is presented. These four levels exhibit distinct differences in typesetting quality: Level-I demonstrates the best typesetting quality, while Level-IV, by contrast, shows the worst.

\begin{figure}[h]
    \centering
    \includegraphics[width=1\linewidth]{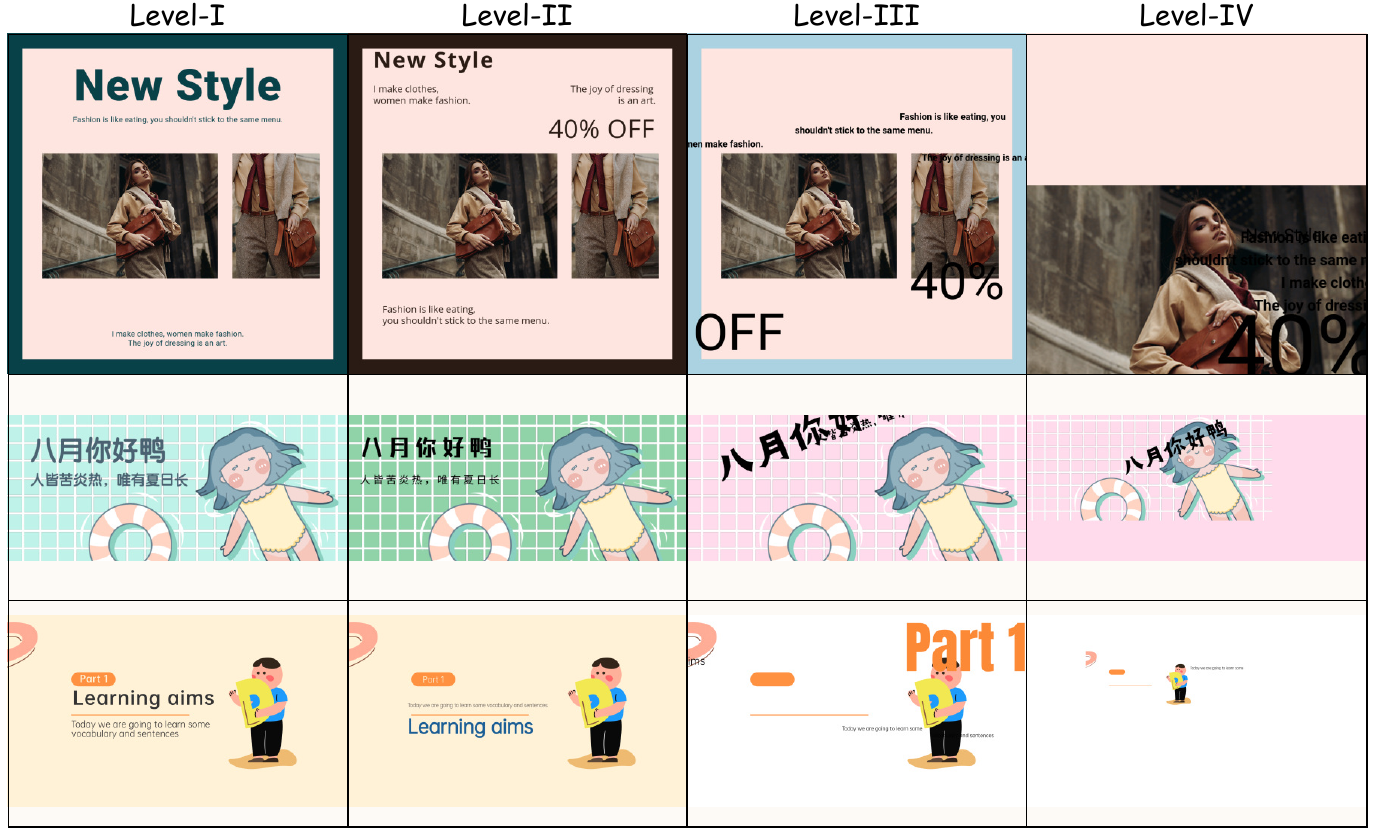}
    \caption{Qualitative comparison of four levels of data in the training data of the reward model}
    \label{fig:rm_data_comparison}
    \vspace*{-\baselineskip}
\end{figure}

\section{Training Details}
\label{sec:train_detail}

\subsection{VFLM}
All experiments are conducted on a cluster of 16 NVIDIA H200 GPUs. For the Cold-Start SFT stage, we trained the model with 8k data for 2 epochs with a learning rate of $1 \times 10^{-5}$ and a batch size of 64. During this stage, the image \texttt{max\_pixels} was set to $1024 \times 28 \times 28$. 

For the RL stage, we set the maximum number of tool calls to 4. The weights for $R_\text{ocr}$ and $R_\text{svg}$ (denoted as $\alpha$) were set to 0.25, while the weight for $R_\text{format}$ was set to 0.1. We prepare up to 32K samples for training, with early stopping based on reward metrics during RL training. We employed a strict on-policy training strategy with the following configuration: batch size of 64, 8 rollouts per sample, sampling temperature of 1.0, KL divergence coefficient of 1e-3, and learning rate of 1e-6.

\subsection{Layout Reward Model}
We trained the reward model on a preference dataset constructed from 200K layout samples.
Four quality levels (Level-I, Level-II, Level-III, Level-IV) were generated for each query, yielding 1.2M preference pairs. We randomly select 25K pairs as the test set, using the remainder for training. During training, we use a batch size of 512 and train for 2100 steps. 

\subsection{Ablation Study}
In the ablation experiments, Single-Round RL, RL from Pretrained, and Direct Output adopt the same hyperparameter configuration as VFLM. For Direct Output SFT, 40k samples are used for SFT, with a batch size of 128, a learning rate of 1e-5, and training conducted for 2 epochs.

\section{Evaluation Details}
\subsection{Evaluation metrics}
We use an OCR engine\footnote{https://github.com/PaddlePaddle/PaddleOCR} to recognize text in design images and evaluate the accuracy of rendered text using character-level f-measure. In the RL reward function, $R_\text{ocr}$ and $R_\text{svg}$ are evaluated using accuracy. Specifically, a character in the OCR recognition result is defined as a True Positive (TP) if it appears in the annotation; otherwise, it is classified as a False Positive (FP). A False Negative (FN) indicates that a character is only present in the annotation but absent from the OCR recognition result. Accordingly, character-level precision, recall, f-measure and accuracy can be formulated as follows:
\begin{equation}
\begin{aligned}
&Char\_P =\frac{TP}{TP+FP}, \\
&Char\_R =\frac{TP}{TP+FN}, \\
&Char\_F =\frac{2\times Char\_P\times Char\_R}{Char\_P+Char\_R}, \\
&Char\_Acc =\frac{TP}{TP+FP+FN}.
\end{aligned}
\label{eq:classification_metrics}
\end{equation}

\begin{figure}[h]
    \centering
    \includegraphics[width=1\linewidth]{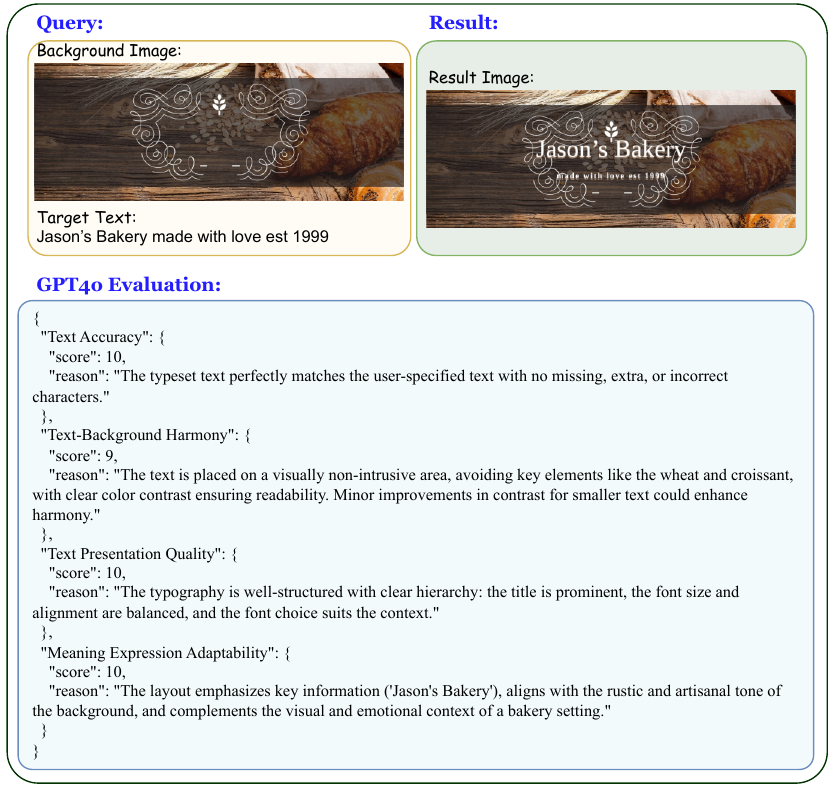}
    \caption{A detailed GPT4o evaluation output.}
    \label{fig:gpt4o_example}
\end{figure}

For the GPT4o evaluation, we assess the effect along four dimensions: Text Accuracy, Text-Background Harmony, Text Presentation Quality, and Meaning Expression Adaptability. The evaluation prompt is shown in prompt~\ref{prompt:GPT4o_Prompt}. \cref{fig:gpt4o_example} is a detailed evaluation sample.

\subsection{Model Prompt}
\subsubsection{VFLM System prompt}
The VFLM system prompt is shown in prompt~\ref{prompt:VFLM_System_Prompt}, includes tool definitions and task descriptions.

\subsubsection{MLLM System prompt}
The system prompt of MLLM models, as shown in prompt~\ref{prompt:MLLM_System_Prompt}, only omits the definition of tools and the statement of multi-round responses compared with that of VFLM.

\section{More Experiments}

\subsection{Number of rounds of reflection}
\begin{figure}[h]
    \centering
    \includegraphics[width=0.8\linewidth]{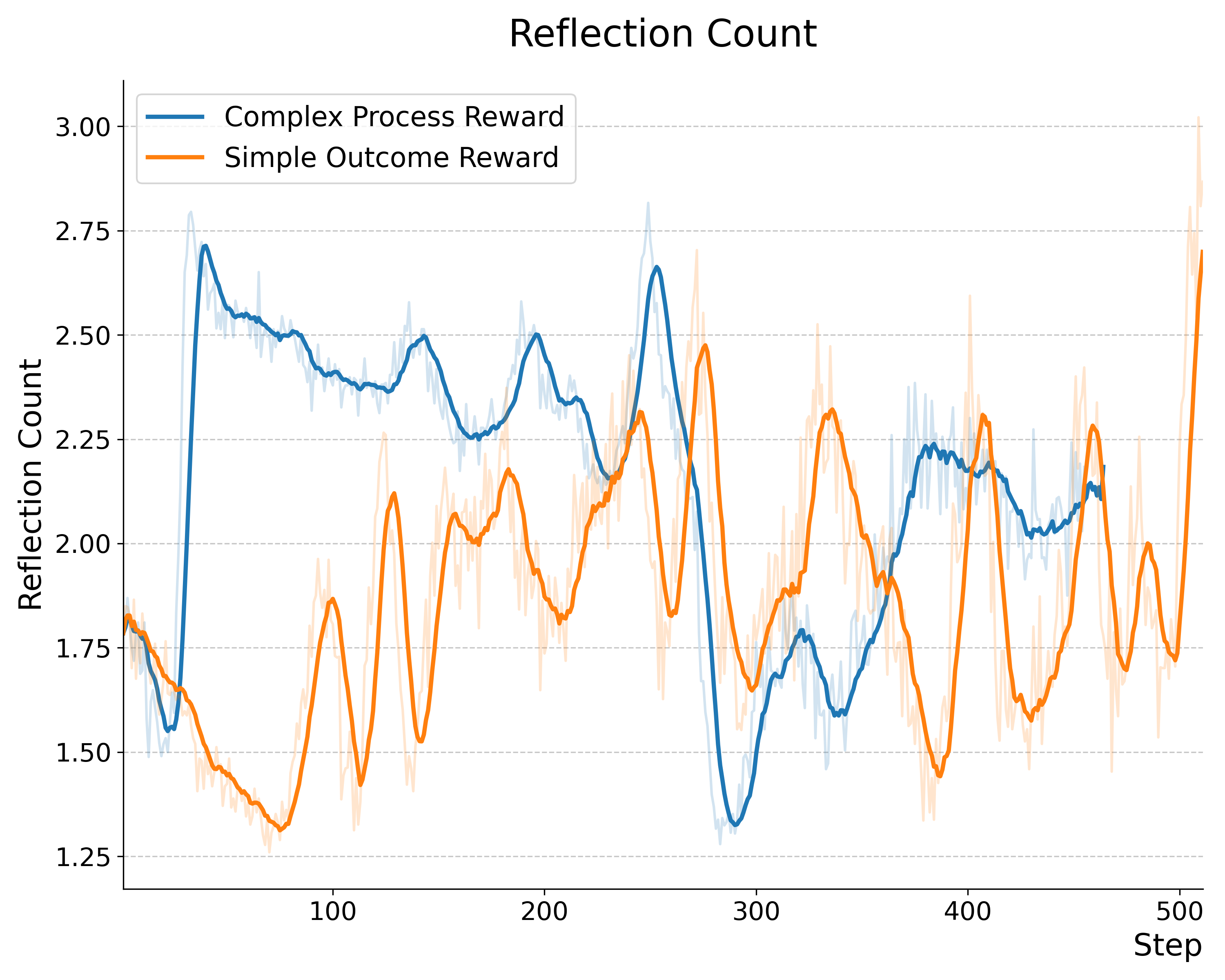}
    \caption{Training processes: reflection times of VFLM, constrained between 1 and 4.}
    \label{fig:train_tool_cnt}
    \vspace{-0.5\baselineskip}
\end{figure}
\cref{fig:train_tool_cnt} illustrates the dynamic evolution of reflection counts during training. While the complex process reward employs strategic termination control (Eq. (9)-Eq. (10)) to enforce stability, VFLM under the simple outcome reward exhibits a distinct, insightful trajectory. Specifically, in the initial phase (the first 100 steps), we observe a decline in reflection turns. Combining with Fig. 4, we can analyze this is attributed to the model's initial instability in output formatting, where iterative attempts often degraded quality compared to the initial generation, prompting the model to curtail its reasoning depth. However, beyond 100 steps, as the output format stabilizes, the model discovers that iterative optimization yields superior rewards. Consequently, the reflection count begins to fluctuate and rise, reflecting the model's autonomous realization that deeper reflection correlates with better layout quality, rather than relying on rigid, pre-defined constraints.

\begin{figure}[h]
    \centering
    \includegraphics[width=\linewidth]{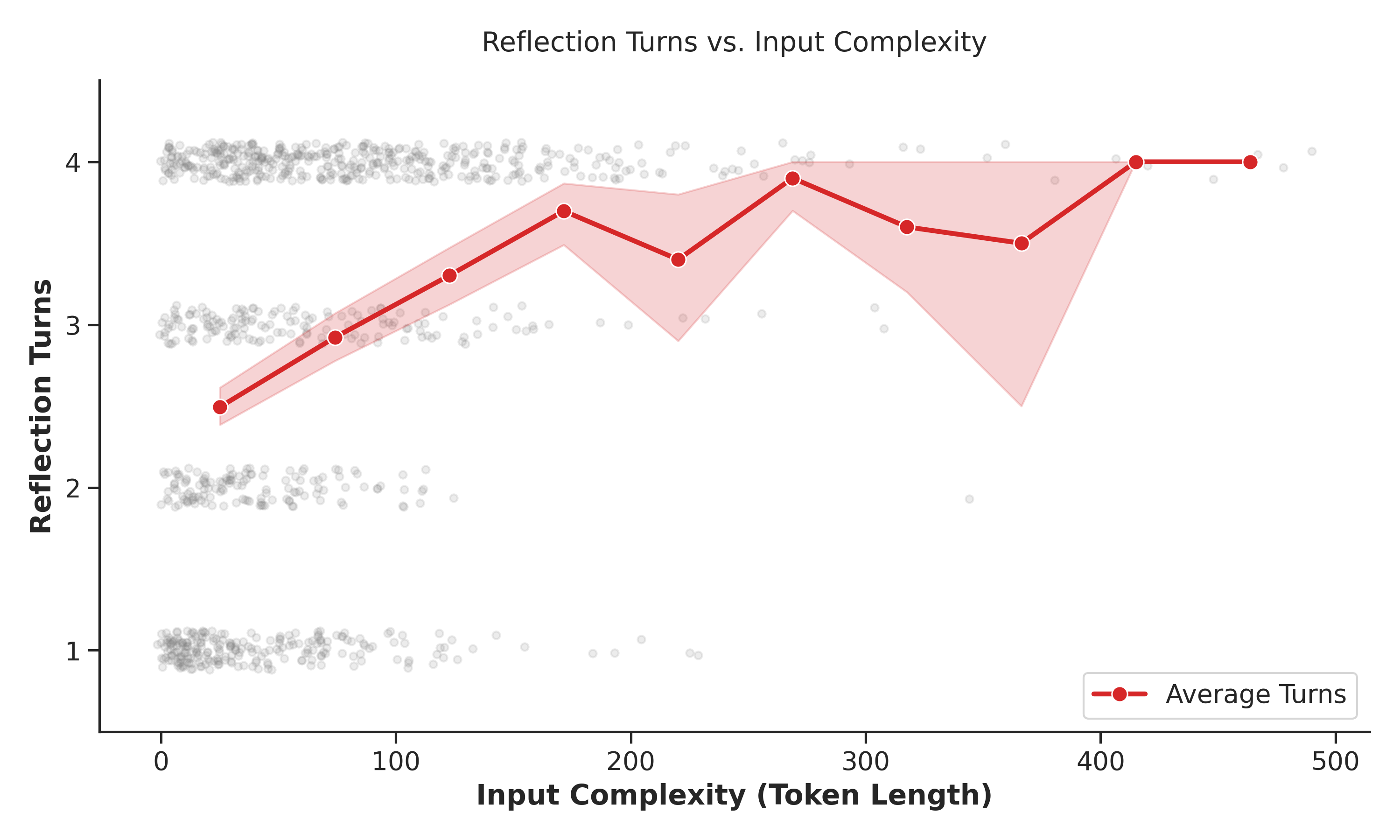}
    \caption{\textbf{Reflection Turns vs. Input Complexity.} The positive trend (red line) demonstrates that VFLM autonomously increases reflection depth for more complex tasks.}
    \label{fig:eval_tool_cnt_text}
    \vspace{-1\baselineskip}
\end{figure}
To further analyze the number of reflection rounds VFLM performs, we examined the correlation between input complexity (measured by the token length of the Target Text) and the number of reasoning rounds. As shown in Figure~\cref{fig:eval_tool_cnt_text}, which presents the model outputs of 1000 samples from VFLM on the TextLayout test set, the gray scatter points depict the raw distribution of inference steps, which are inherently discrete integers. To visualize the underlying trend amidst this variance, the red solid line tracks the average reflection turns across complexity intervals. A clear positive trend is observable: for concise inputs ($<100$ tokens), the model efficiently converges with fewer refinement steps (averaging $\sim$2.5 turns). In contrast, as input complexity increases to over 400 tokens, the model adaptively increases its reasoning depth, approaching the maximum of 4 turns. This confirms that VFLM actively perceives layout difficulty and allocates computational resources accordingly.

\subsection{Human Study}
\begin{figure}[!h]
\vspace{-\baselineskip}
    \centering
    \includegraphics[width=\linewidth]{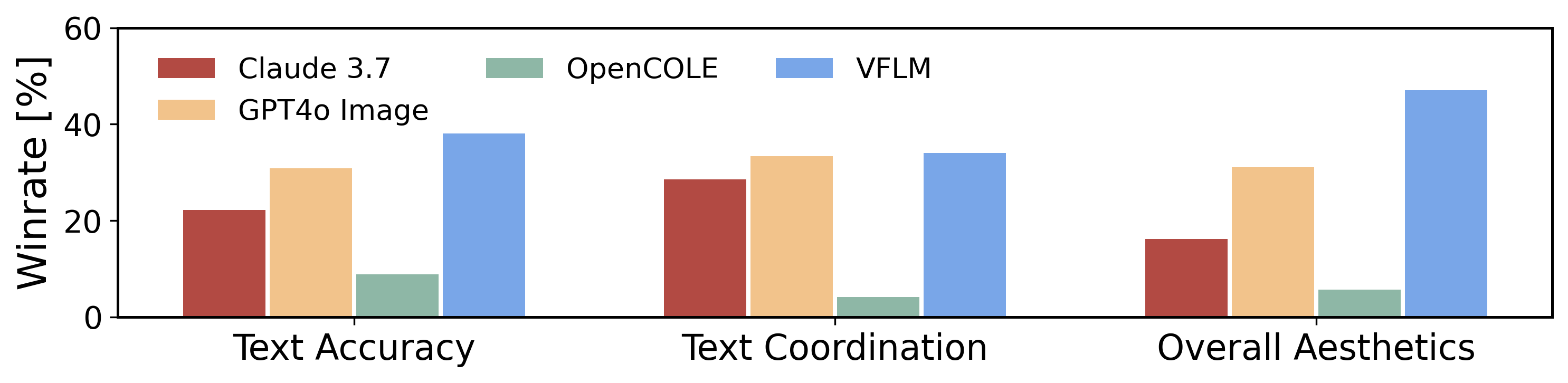}
    \caption{Results of human study.}
    \label{fig:human_study}
\end{figure}

To further enhance evaluation reliability, we supplement our analysis with a blind human study involving 16 participants, who evaluated outputs from Claude3.7, GPT4o-Image, OpenCOLE, and VFLM on three dimensions across 320 randomly selected queries from three datasets. A total of 960 votes were collected. As shown in the \Cref{fig:human_study}, VFLM performed comparably to GPT4o-Image on text coordination and achieved the best results in text accuracy and overall aesthetics.

\subsection{Compute Cost and Latency}
\begin{figure}[!h]
    \centering
    \includegraphics[width=1\linewidth]{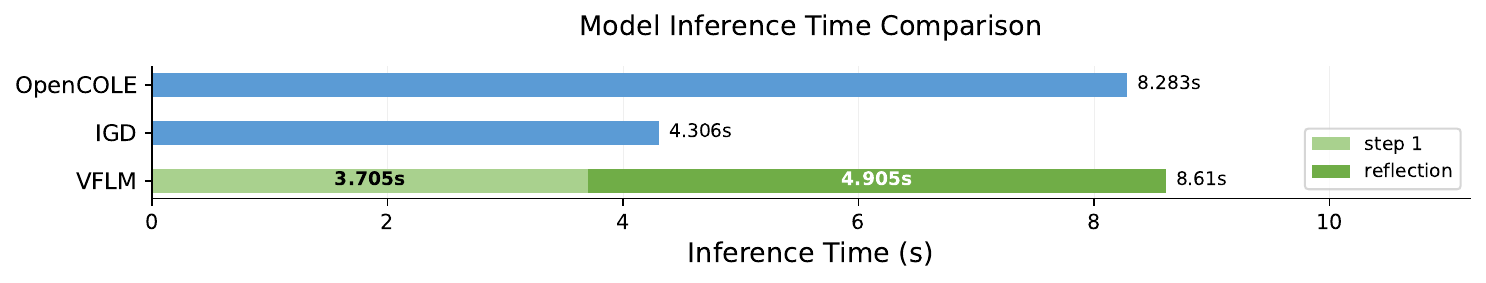}
    \caption{Comparison of inference time and latency with OpenCOLE and IGD.}
    \label{fig:infer_time}
\end{figure}

As shown in \Cref{fig:infer_time}, we report the average inference time of VFLM, OpenCOLE, and IGD on the TextLayout test set using 4×4090 GPUs. Although VFLM has the highest latency, it yields substantial performance gains.

\subsection{More Baseline}
\begin{table}[htbp]
\centering
\caption{Metrics and RM score on OpenCOLE* and MLLM feedback baseline.}
\label{tab:more_baseline}   
\resizebox{\linewidth}{!}
{
\begin{tabular}{c|c|ccc|c}
\hline
Model                                & Char-F $\uparrow$ & $R_{ali} \downarrow$ & $R_{ove} \downarrow$ & $R_{com} \downarrow$ & RM Score $\uparrow$ \\ \hline
OpenCOLE*                            & 0.7671            & 0.7114               & 0.0143               & 21.8279              & 0.3620              \\
VFLM                                 & \textbf{0.9071}            & 0.0035               & \textbf{0.0059}               & \textbf{15.4583}              & \textbf{0.5415}              \\
Qwen2.5-VL-7B-feedback   & 0.6830$\blacktriangle$            & \textbf{0.0025}$\blacktriangle$               & 0.0188$\blacktriangle$               & 28.8221$\blacktriangledown$              & 0.0910$\blacktriangledown$              \\
GPT-4o-feedback & 0.8494$\blacktriangle$            & 0.0039$\blacktriangle$               & 0.0028$\blacktriangle$               & 20.8852$\blacktriangledown$              & 0.3662$\blacktriangle$             \\ \hline
\end{tabular}
}
\end{table}

The table above shows the results of OpenCOLE trained on the same training set as VFLM. Although retraining brings substantial improvements over the original OpenCOLE, it still underperforms VFLM. We further integrate visual feedback into GPT-4o and Qwen2.5-VL-7B baselines. As shown in the table, without task-specific training, introducing visual feedback to MLLMs yields only marginal gains, with some metrics even showing slight declines.

\subsection{Ablation on Other Datasets}
\begin{table}[htbp]
\centering
\caption{Ablation experiments on Graphic quality metrics and OCR metrics on Crello test set.}  
\label{tab:ablation_crello}   
\resizebox{\linewidth}{!}{
\begin{tabular}{c|c|ccc|c}
\hline
\multirow{2}{*}{Model} & OCR               & \multicolumn{3}{c|}{Graphic}                                       & \multicolumn{1}{l}{\multirow{2}{*}{RM Score $\uparrow$}} \\
                       & Char-F $\uparrow$ & $R_{ali} \downarrow$ & $R_{ove} \downarrow$ & $R_{com} \downarrow$ & \multicolumn{1}{l}{}                                     \\ \hline
VFLM-step1             & 0.8774            & 0.0046               & 0.0061               & 19.5917              & 0.4392                                                   \\
VFLM                   & \textbf{0.9256}   & 0.0025               & {\ul 0.0022}         & \textbf{14.8063}     & \textbf{0.5548}                                          \\ \hline
Cold-Start-step1       & 0.7904            & 0.0118               & 0.0410               & 24.3143              & 0.2721                                                   \\
Cold-Start             & 0.7928            & 0.0121               & 0.0402               & 24.0888              & 0.2774                                                   \\ \hline
Single-Round RL        & 0.8829            & 0.0010               & 0.0094               & 24.4845              & 0.4007                                                   \\
RL from Pretrained     & 0.8792            & \textbf{0.0004}      & \textbf{0.0012}      & 30.3397              & 0.3482                                                   \\
Direct Output          & {\ul 0.9092}      & {\ul 0.0009}         & {\ul 0.0022}         & 19.3983              & 0.4596                                                   \\
Direct Output SFT      & 0.8960            & 0.0032               & 0.0192               & {\ul 18.3754}        & {\ul 0.4680}                                             \\ \hline
\end{tabular}
}
\end{table}
\begin{table}[htbp]
\centering
\caption{Ablation experiments on Graphic quality metrics and OCR metrics on DESIGNERINTENTION test set.}  
\label{tab:ablation_cole}   
\resizebox{\linewidth}{!}{
\begin{tabular}{c|c|ccc|c}
\hline
\multirow{2}{*}{Model} & OCR               & \multicolumn{3}{c|}{Graphic}                                       & \multicolumn{1}{l}{\multirow{2}{*}{RM Score $\uparrow$}} \\
                       & Char-F $\uparrow$ & $R_{ali} \downarrow$ & $R_{ove} \downarrow$ & $R_{com} \downarrow$ & \multicolumn{1}{l}{}                                     \\ \hline
VFLM-step1             & 0.9415            & 0.0033               & 0.0023               & 14.1230              & 0.5285                                                   \\
VFLM                   & \textbf{0.9663}   & 0.0024               & {\ul 0.0008}         & \textbf{12.1167}     & \textbf{0.5688}                                          \\ \hline
Cold-Start-step1       & 0.8860            & 0.0130               & 0.0264               & 17.5229              & 0.3784                                                   \\
Cold-Start             & 0.8881            & 0.0127               & 0.0257               & 16.9280              & 0.3850                                                   \\ \hline
Single-Round RL        & 0.9382            & {\ul 0.0006}         & 0.0033               & 16.7466              & 0.4768                                                   \\
RL from Pretrained     & 0.9430            & \textbf{0.0005}      & \textbf{0.0001}      & 18.0272              & 0.4692                                                   \\
Direct Output          & {\ul 0.9590}      & \textbf{0.0005}      & 0.0010               & 14.6338              & 0.5167                                                   \\
Direct Output SFT      & 0.9434            & 0.0017               & 0.0082               & {\ul 13.1702}        & {\ul 0.5398}                                             \\ \hline
\end{tabular}
}
\end{table}

\cref{tab:ablation_crello} and \cref{tab:ablation_cole} demonstrate the metrics of all models in the ablation experiments on the Crello and DESIGNERINTENTION test sets. Consistent with the conclusions in the TextLayout dataset, VFLM also achieves significant advantages over other baselines on these two datasets, fully demonstrating the generalization ability of our visual feedback method.

\subsection{Ablation on Reward Functions}

In our RL training process, the score for layout performance ($R_\text{score}$) from the reward model and two rewards based on text accuracy:
\begin{equation}
    R_\text{score} = R_{\text{layout}} + \alpha \cdot \left( R_{\text{ocr}} + R_{\text{svg}}\right).
\end{equation}
We perform ablation to investigate whether using only $R_\text{score}$ from the reward model is effective. To save validation time, tests are conducted on the models of the two ablation experiments: Single-Round RL and Direct Output.

\begin{table}[htbp]
\centering
\caption{Ablation experiments on the reward function on the TextLayout test set.}  
\label{tab:ablation_reward_fun}   
\resizebox{\linewidth}{!}{
\begin{tabular}{c|c|ccc|c}
\hline
\multirow{2}{*}{Model}  & OCR               & \multicolumn{3}{c|}{Graphic}                                       & \multicolumn{1}{l}{\multirow{2}{*}{RM Score $\uparrow$}} \\
                        & Char-F $\uparrow$ & $R_{ali} \downarrow$ & $R_{ove} \downarrow$ & $R_{com} \downarrow$ & \multicolumn{1}{l}{}                                     \\ \hline
Single-Round RL         & \textbf{0.8792}   & 0.0024               & \textbf{0.0053}      & \textbf{18.9428}     & 0.4063                                                   \\
Single-Round RL-w/o OCR & 0.8696            & 0.0024               & 0.0054               & 20.0621              & \textbf{0.4153}                                          \\ \hline
Direct Output           & \textbf{0.9237}   & 0.0027               & \textbf{0.0021}      & 17.0654              & 0.4964                                                   \\
Direct Output-w/vo OCR   & 0.9146            & \textbf{0.0026}      & 0.0059               & \textbf{16.6961}     & \textbf{0.5039}                                          \\ \hline
\end{tabular}
}
\end{table}

The results are presented in \cref{tab:ablation_reward_fun}. It can be observed from the table that the two models exhibit little difference in RL training performance with and without the $R_\text{ocr}$ and $R_\text{svg}$ rewards, with many metrics being relatively close. Since 
$R_\text{ocr}$ and $R_\text{svg}$ have low computational costs and the task essentially involves text typesetting, these two reward functions are retained in the final training of VFLM.

\section{Case Study}
\begin{figure*}
    \centering
    \includegraphics[width=0.9\linewidth]{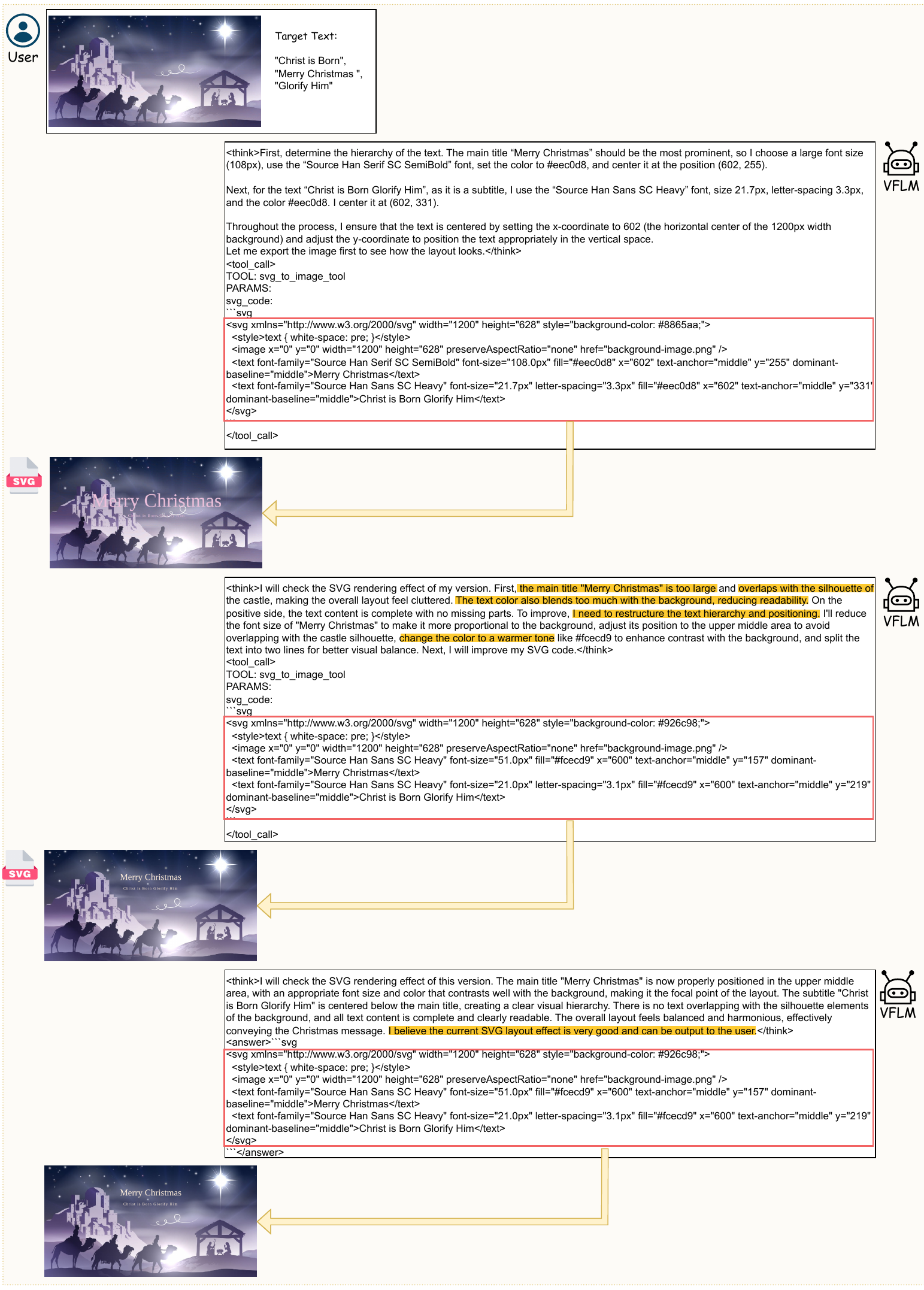}
    \caption{A detailed VFLM output case.} 
    \label{fig:a_case}
\end{figure*}

\begin{figure*}
    \centering
    \includegraphics[width=0.8\linewidth]{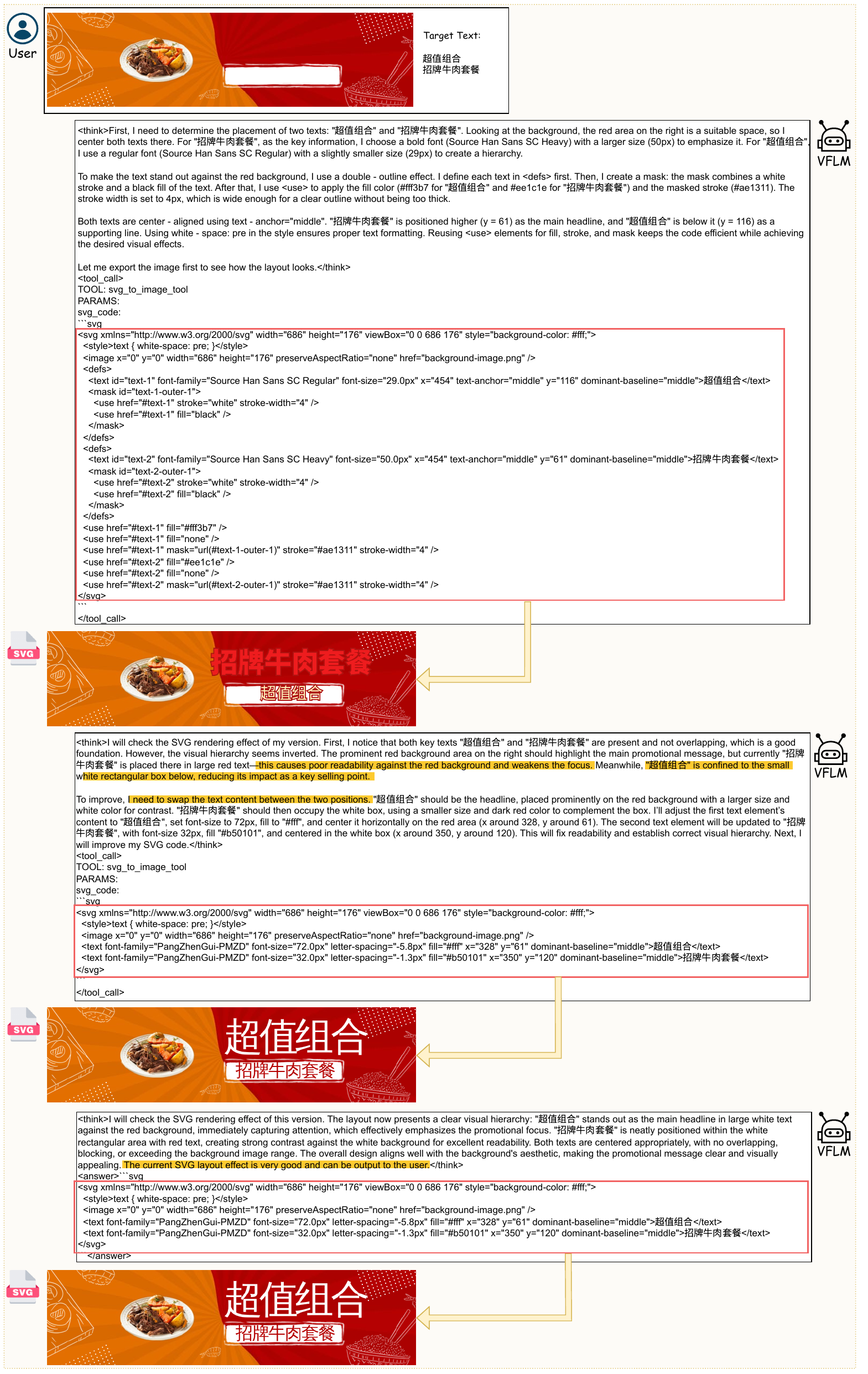}
    \caption{Another detailed VFLM output case.} 
    \label{fig:another_case}
\end{figure*}

\cref{fig:a_case} presents a detailed case generated by VFLM. The SVG initially produced by VFLM exhibits several obvious issues in terms of rendering effect; for instance, the main title "Merry Christmas" is too large and overlaps with the silhouette of the castle. After receiving visual feedback from the rendered image, it can be observed from the model's reasoning content (highlighted in the figure) that the model identified these problems and implemented appropriate improvements. Following the second round of visual feedback, the model deemed the current layout satisfactory and output it to user as the final result. \cref{fig:another_case} is another case.

\begin{figure*}
    \centering
    \includegraphics[width=\linewidth]{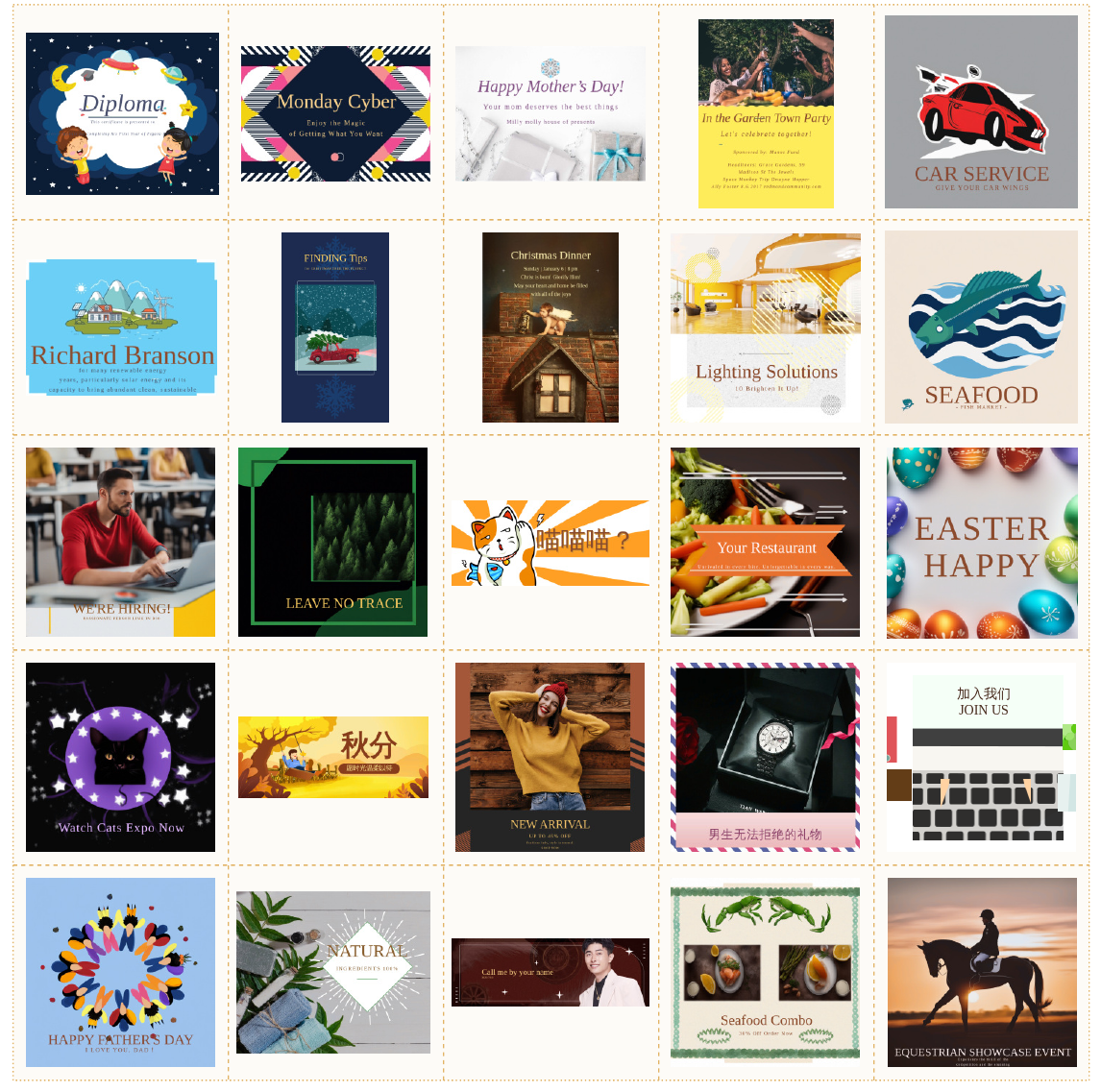}
    \caption{More VFLM output cases.} 
    \label{fig:show_case}
\end{figure*}

\begin{figure*}
    \centering
    \includegraphics[width=\linewidth]{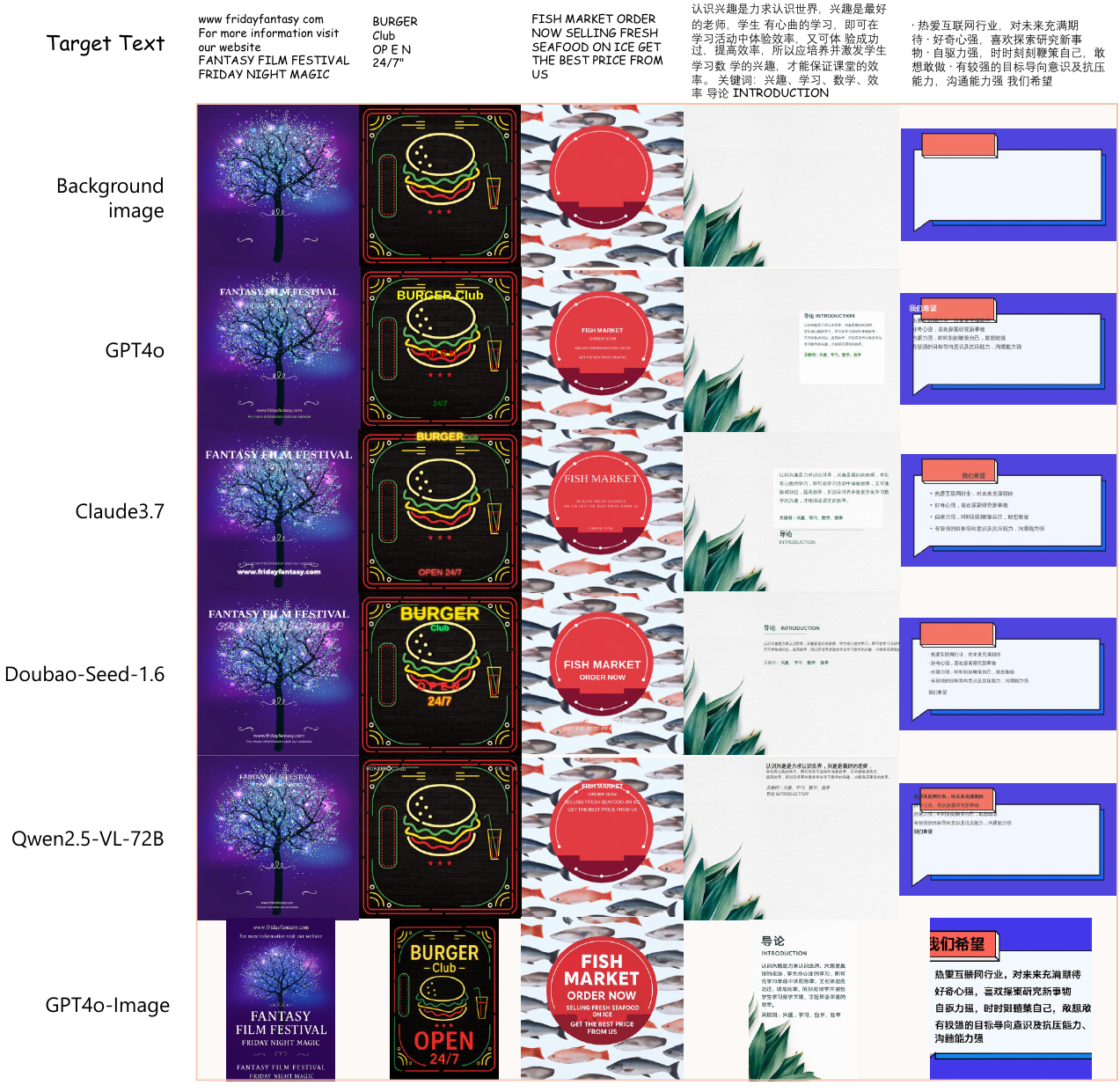}
    \caption{In comparison with all existing methods(Part1).}
    \label{fig:pic_full_1}
\end{figure*}

\begin{figure*}
    \centering
    \includegraphics[width=\linewidth]{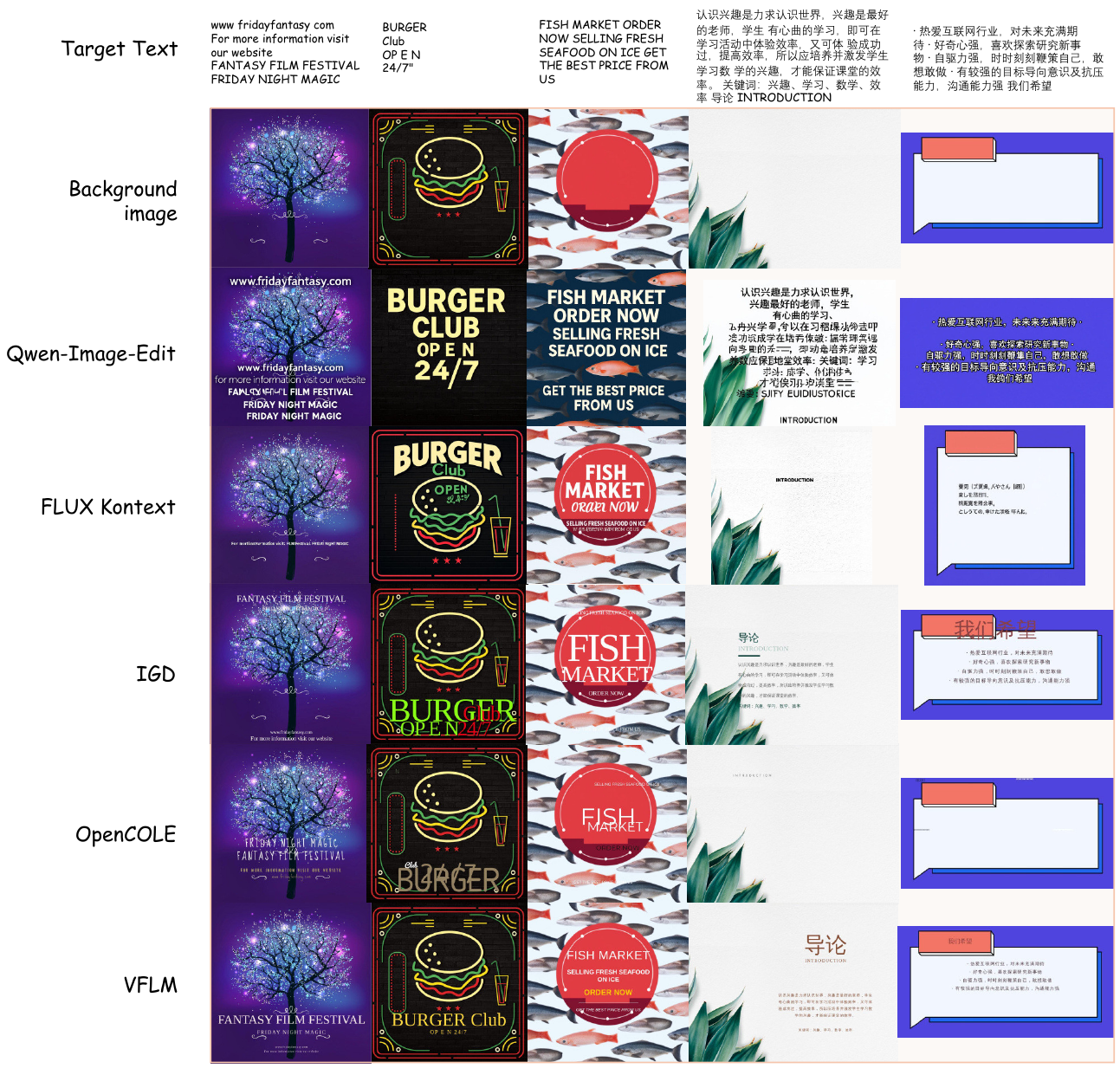}
    \caption{In comparison with all existing methods(Part2).}
    \label{fig:pic_full_2}
\end{figure*}

\cref{fig:show_case} show the more output cases of VFLM. \cref{fig:pic_full_1} and \cref{fig:pic_full_2} demonstrate the comparison between VFLM and all other models.

\begin{promptbox*}[Initial Reasoning Process Prompt]{lightblue}{prompt:Initial_Reasoning}
\begin{Verbatim}[breaklines,breakanywhere,breaksymbol={}]
Role setting:
You are an experienced Layout and SVG engineer.

Task:
Here is a result of using SVG code to typeset specific text on an input background image. I will provide you with the designed SVG code and the rendered image of this code, which has a very beautiful layout effect.
Now, assuming you are the designer of typesetting this SVG, what is your thought process when typesetting this SVG?
Could you please use the voice of a designer to briefly describe your thought process when designing this SVG based on the SVG code and rendering results? How did you design this SVG?
Ensure that your design ideas are consistent and closely related to the design results of this SVG. Do not fabricate content that is not included in the SVG, as the SVG only typesets the given text based on the given background image. Therefore, the typesetting ideas should focus on the implementation of the text, including text position, font, size, beautification effect, etc. Don't pay attention to the text already in the background image. Do not pay attention to unnecessary other graphic icons and other elements, and do not analyze how good the rendered image is from the perspective of typesetting, because your thinking is based on not seeing the rendered image.

background-image.png: <image>
image size : {image_width} * {image_height}
texts: {target_text}

svg code:
```svg
{svg_code}
```
The image result rendered by this svg: <image>

Output requirements:
Please directly output the thinking process of the person in the designer's tone, without any other descriptive content. Answer in English.
\end{Verbatim}
\end{promptbox*}

\begin{promptbox*}[Multi-Round Reflection Data Synthesis Prompt]{lightblue}{prompt:Multi_round_reflection}
\begin{Verbatim}[breaklines,breaksymbol={}]
USER:

You are an experienced typesetter and SVG engineer, skilled at elegantly typesetting specified text on user provided background images.
You know how to apply unique aesthetic principles to design professional and attractive layout, using SVG code to create beautiful layouts. Please design the final layout plan based on the background image and text content provided by the user.
In SVG code, use the image tag to reference the background image: href="background-image.png", while other elements only need to be designed with content related to the text.
Please design an SVG code layout scheme based on the background image and text content provided by the following users. 
You first think about it, and then output the final SVG code. The format is<think>...</think>\n<answer>...</answer>

background-image.png: <image>
image size: {image_width}*{image_height}
Please arrange the following text on the background image: 
{target_text}

ASSISTANT:
{Reasoning Process and SVG generated by Qwen2.5-VL-7B}

USER:
Now, I'll show you the effect of this version of SVG layout, and you need to improve this SVG layout effect.
I'll also give you a standard SVG layout result, and you need to improve your SVG layout according to this standard SVG layout result.

- In your output, you need to speak in the tone of a designer, stating that you've reviewed the SVG result of your initial layout, then reflected on it and made corrections. Note that you've designed an initial version of the SVG, and now I've provided you with the rendered image. Your output should focus on examining the image, ensuring it's a reflection and correction of your initial SVG layout result. The direction of correction is the correct effect I gave you, but don't expose in the output that you're improving based on the standard effect. Pretend you've thought it out on your own.

- The output should include your thinking process for SVG layout, how to improve your SVG layout result step by step. You need to point out which parts of your initial layout were good and which were bad and needed modification. For each modification point, be specific about how to modify the SVG code. Don't just qualitatively say which aspects you'll modify. Pay attention to the tone, which should be like that of a designer, and the content of the output should conform to the designer's way of thinking.

- During the modification process, key considerations should be text position, whether there is any text missing, text overlapping, text being blocked, and whether the text exceeds the background image range, etc. These considerations need to be included in the output.

- Your output modification process may involve multiple steps. If your initial layout is not very different from the standard one, you can make only one modification; if there is a large gap, multiple steps of modification are required. You need to simulate the designer's thinking process and gradually improve the SVG layout. Each time you modify, choose the part with the worst effect to improve. Explain the specific SVG code improvements in the thinking process. After modifying one version, only make changes to the SVG part that needs to be modified in this step, and don't change the other parts for now. Output the complete SVG code; then proceed to the next modification until you think the SVG layout effect is very good. Don't make too many modifications. Ensure that each modification is better than the previous one, with a maximum of 3 modifications. The SVG code after the last modification needs to be output, and its effect should be the same as that of the standard code I gave you.

- You need to answer one modification each time, and then I'll show you the rendered effect of the SVG you modified, and you'll make the next modification.

- Based on the rendered image effect I give you after each of your modifications, decide whether the next modification is needed. Each modification should have a significant improvement, not just a minor one. For example, when the order of different text tags doesn't affect the SVG rendering effect, there's no need for additional modification. Since I require you to make as few steps of modification as possible, each modification should have a significant improvement.

Your output is the thinking process of a designer improving the SVG layout after reviewing the first version they designed. I've given you the standard SVG code, and you should modify the SVG code in this direction. However, note that your output is based on not having seen this standard SVG, as if the designer is reflecting after designing the initial draft and modifying it to the final standard SVG version through multiple steps.

After the last modification, you need to output the final inspection, indicating that after checking the image, you think the current SVG layout effect is very good and can be replied to the user.

Your initial SVG layout effect is shown in the figure below:
<image>
This is standard and beautiful SVG code. The code and its rendered effect diagram are as follows.
```svg
{svg_code}
```
<image>

Output requirements:
Please directly output the thinking process in the tone of a designer, without any other descriptive content. Be careful not to reveal that you have seen the standard SVG effect. Transform it into your own thinking. The output should conform to the designer's thinking process, that is, how you think about improving the layout by yourself, not by comparing with the standard effect. Do not output the word "standard".

If improvement is needed, the first sentence in each step of the thinking process should be: "I will check the SVG rendering effect of my version...", and the last sentence should be: "Next, I will improve my SVG code."

These two beginning and ending sentences are necessary and cannot be omitted, but you can modify the language to maintain the same meaning and make the output diverse.

Your output needs to specifically point out which effects in your first version are good and do not need improvement, which effects are poor and need improvement, and how to specifically modify the SVG code. If you think the SVG layout effect of your first version is very close to or even better than the standard SVG rendering effect I provided, you can describe your satisfaction with this SVG layout and that you think it has achieved a very good effect and does not need further improvement.

Answer in English.

Output requirements:
- You need to output in the form of multi - round conversations. According to the number of modifications you decide, the output format for each modification is as follows:

# Step {current modification number} of modification:

## Thinking process: Here, think about how to make the modification.

## SVG code: Modify the complete SVG code.

- After the final modification, the rendering result of your SVG code should be exactly the same as that of the reference SVG code I provided.

- After the last modification is output, I will provide you with the rendered image again. Then you need to output a final reflection, indicating that you will check the SVG rendering effect of this version and think that the current SVG layout effect is very good and does not need to be improved further, and it can be output to the user. The output format of the final reflection is:

# Final rethink: ...

USER:
Your current SVG layout effect is shown in the figure below:
<image>

ASSISTANT:
...
\end{Verbatim}
\end{promptbox*}

\begin{promptbox*}[VFLM System Prompt]{lightblue}{prompt:VFLM_System_Prompt}
\begin{Verbatim}[breaklines,breakanywhere,breaksymbol={}]
System:
You are a helpful assistant.

# Tools
You may call one or more functions to assist with the user query.

You are provided with function signatures within <tools></tools> XML tags:
<tools>
{
    "type": "function ",
    "function": {
        "name": "svg_to_image_tool",
        "description": Convert SVG code to an image.",
        "parameters": {
            "type": "object",
            "properties": {
                "svg_code": {
                    "type": "string",
                    "description": "The SVG code to convert to an image."
                }
            },
            "required":[
                "svg_code"
            ]
        }
    }
}
</tools>

# How to call this tool
Wrap the SVG code with specific markers (``` and ```) within <tool_call></tool_call> XML tags.

**Example**: 
<tool_call>
TOOL: svg_to_image_tool
PARAMS: 
svg_code:
```svg
...
```
</tool_call>

You are an experienced visual layout designer and SVG engineer, skilled at elegantly typesetting specified text on background images provided by users.
You know how to apply unique aesthetic principles to design professional and appealing layouts, using SVG code to create beautiful layouts. Please design a final layout plan based on the background image and text content provided by the user.
In the SVG code, use the image tag to reference the background image: href=\"background-image.png\", and other elements only need to design content related to the text.
Please design an SVG code layout plan based on the following background image and text content provided by the user.
You should first view the background image, think about how to typeset the text on the background image, design a version of SVG code, correctly reference the background image in the SVG code, then call the svg_to_image tool, and you will get the picture of your SVG. Then, based on the picture, judge whether the typesetting of your picture meets the expectations, whether the background image is correctly referenced, and whether the text is beautiful. If the typesetting effect is not good enough, modify the SVG code, and repeatedly reflect after tool calls until the typesetting effect is better. Finally, output the final SVG code.
Format: <think>...</think>\n<tool_call>...</tool_call>(if tools needed) <answer>...</answer>

User:
background-image.png: <image>
image size: {image_width}*{image_height}
texts: {text}
\end{Verbatim}
\end{promptbox*}

\begin{promptbox*}[MLLM System Prompt]{lightblue}{prompt:MLLM_System_Prompt}
\begin{Verbatim}[breaklines,breakanywhere,breaksymbol={}]
System:
You are an experienced layout designer and SVG engineer, proficient in elegantly laying out specified text on a background image provided by the user. You have a deep understanding of how to use unique aesthetic principles to design a professional and attractive layout. Use SVG code to create a beautiful layout. Please design the final layout plan according to the background image and text content provided by the user. In the SVG code, use the image tag to reference the background image: href="background-image.png", and only design the elements related to the text. Please design the SVG code layout plan according to the background image and text content provided below.

User:
background-image.png: <image>
image size: {image_width}*{image_height}
texts: {text}
\end{Verbatim}
\end{promptbox*}

\begin{promptbox*}[GPT4o Evaluation Prompt]{lightblue}{prompt:GPT4o_Prompt}
\begin{Verbatim}[breaklines,breakanywhere,breaksymbol={}]
You are an autonomous AI Assistant specializing in evaluating the typesetting effects of a typesetting model. This model's core task is to typeset user-specified text on a background image; your goal is to provide objective, targeted, and constructive scoring and feedback based on text-typesetting-specific principles and practical application needs. Your evaluation covers four independent dimensions: text content accuracy, text-background visual harmony, text presentation quality, and meaning expression adaptability. You will be provided with the background image, the user's original specified text, and the typeset result (background image + typeset text). Your task is to score the typesetting effect objectively based on the following 4 criteria and provide concise reasoning for each score.

Scoring rules:
- For each of the 4 criteria, score objectively and rigorously on an independent scale of 1-10. For a single criterion, a score of 10 means flawless performance (no issues, fully meeting expectations); a score of 7 indicates minor flaws (no impact on core performance); a score of 4 reflects significant shortcomings (affecting core performance); a score of 1-2 signifies severe issues (rendering the function of this criterion ineffective).
- Keep reasoning concise (1-2 sentences per criterion), focusing on specific performance. If the output is too long, it will be truncated.
- Only respond in JSON format with 4 top-level keys corresponding to the 4 Grading criteria. Each key's value is an object containing "score" (integer 1-10) and "reason" (string). No other information.

Grading criteria:
1. Text Accuracy (1-10): Evaluate consistency with the user's original text (no missing/extra/wrong characters, no spelling/grammatical errors in Chinese/English). Score 10: 100% accurate; Score 1: massive errors or unrecognizable characters.

2. Text-Background Harmony (1-10): Evaluate visual coordination: (1) text avoids blocking the background's main subject (key figures, core graphics); (2) text color/transparency ensures clear contrast with the background (no blurring). Score 10: no blocking, perfect contrast; Score 1: complete blocking or unreadable due to poor contrast.

3. Text Presentation Quality (1-10): Evaluate text's own properties: (1) structural rationality (clear title/body hierarchy, compliance with reading habits, balanced spacing); (2) physical readability (appropriate font selection, suitable size, neat alignment). Score 10: clear structure, highly readable; Score 1: chaotic structure and physically unreadable.

4. Meaning Expression Adaptability (1-10): Evaluate meaning transmission: (1) key information is highlighted (via weight/color/size); (2) layout matches text's emotional tone (e.g., serious text uses rigorous typography); (3) text position aligns with the background's semantic context (e.g., "ocean protection" text near ocean elements). Score 10: amplifies meaning, matches tone, aligns with background semantics; Score 1: contradicts meaning/tone or conflicts with background semantics.


The background image:
<image>
User's original specified text: {text_content}

The typeset result:
<image>
\end{Verbatim}
\end{promptbox*}

\end{document}